\documentclass{article}

\usepackage{microtype}
\usepackage{graphicx}
\usepackage{subcaption}
\usepackage{booktabs} % for professional tables

\usepackage{multirow}

\usepackage{hyperref}

\usepackage[accepted]{icml2023}

\usepackage{amsmath}
\usepackage{amssymb}
\usepackage{mathtools}
\usepackage{amsthm}
\usepackage{bbold}
\usepackage{bbm}
\usepackage{bm}

\usepackage[capitalize,noabbrev]{cleveref}

\theoremstyle{plain}
\newtheorem{theorem}{Theorem}[section]

\newtheorem{corollary}[theorem]{Corollary}
\theoremstyle{definition}
\newtheorem{definition}[theorem]{Definition}
\newtheorem{assumption}[theorem]{Assumption}
\theoremstyle{remark}
\newtheorem{remark}[theorem]{Remark}

\usepackage[textsize=tiny]{todonotes}

\usepackage[mathscr]{euscript}

\renewcommand{\algorithmiccomment}[1]{#1}

\icmltitlerunning{Federated Conformal Predictors for Distributed Uncertainty Quantification}

\begin{document}

\twocolumn[
\icmltitle{Federated Conformal Predictors for Distributed Uncertainty Quantification}



\icmlsetsymbol{equal}{*}

\begin{icmlauthorlist}
\icmlauthor{Charles Lu}{mit,equal}
\icmlauthor{Yaodong Yu}{berk,equal}
\icmlauthor{Sai Praneeth Karimireddy}{berk}
\icmlauthor{Michael I. Jordan}{berk}
\icmlauthor{Ramesh Raskar}{mit}

\end{icmlauthorlist}
\icmlaffiliation{mit}{MIT Media Lab, Massachusetts Institute of Technology, Cambridge, USA}
\icmlaffiliation{berk}{Department of Electrical Engineering and Computer Sciences, University of California, Berkeley, Berkeley, USA}

\icmlcorrespondingauthor{Charles Lu}{luchar@mit.edu}
\icmlcorrespondingauthor{Yaodong Yu}{yyu@eecs.berkeley.edu}

\icmlkeywords{Federated Learning, Conformal Prediction}

\vskip 0.3in
]



\printAffiliationsAndNotice{\icmlEqualContribution} % otherwise use the standard text.

\begin{abstract}
    Conformal prediction is emerging as a popular paradigm for providing rigorous uncertainty quantification in machine learning since it can be easily applied as a post-processing step to already trained models.
    In this paper, we extend conformal prediction to the federated learning setting. 
    The main challenge we face is data heterogeneity across the clients --- this violates the fundamental tenet of \emph{exchangeability} required for conformal prediction. 
    We propose a weaker notion of \emph{partial exchangeability}, better suited to the FL setting, and use it to develop the Federated Conformal Prediction (FCP) framework. 
    We show FCP enjoys rigorous theoretical guarantees and excellent empirical performance on several computer vision and medical imaging datasets.
    Our results demonstrate a practical approach to incorporating meaningful uncertainty quantification in distributed and heterogeneous environments.
    We provide code used in our experiments \url{https://github.com/clu5/federated-conformal}.
\end{abstract}

\section{Introduction}
\label{sec:intro}
For many real-world machine learning applications, predictive performance should not be the sole criterion determining a model's usefulness~\cite{bansal2019beyond,babbar2022utility}.
For example, with AI medical devices, predictive performance should be considered in the context of principled uncertainty quantification~\cite{kompa2021second}.
Techniques that provide meaningful estimates of uncertainty quantification, such as conformal prediction~\cite{vovk2005algorithmic}, are critical to deploying machine learning in safety-conscious domains such as healthcare~\cite{bhatt2021uncertainty}.

Consider how a doctor performs differential diagnosis on a patient by leveraging their clinical training and intuition to rule out highly unlikely conditions to produce a narrow list of conditions for a particular patient with some observed symptoms. 
Thus, instead of a point estimate, we end up with a \emph{prediction set} of possible conditions.
The number of conditions in this prediction set naturally expresses uncertainty --- a large set implies high uncertainty, while a small set implies low uncertainty.
Using such a prediction set instead of a single prediction is more interpretable. It can increase trust in black-box models for safety-critical decision-making~\cite{lu2022improving,lu2022fair} and has been promoted for several medical applications~\cite{shashikumar2021artificial,vazquez2022conformal,angelopoulos2022image,lu2022three}.

Furthermore, \textit{conformal prediction} insists that these prediction sets should contain the correct output with high probability.
Concretely, if $\mathcal{C}(X)$ is some set-valued function that generates prediction sets on $X$ and $\alpha \in (0, 1)$ is the desired confidence level, we would like the constructed prediction set to contain the true output with probability $\mathbf{P}\left(Y \in \, \mathcal{C}(X)\right) \geq 1 - \alpha.$

In this paper, we consider conformal prediction in the federated learning (FL) setting, where several clients (each with some local data distribution $\mathbb{P}_{k}$) jointly optimize a shared global model on a global distribution. 
Our goal is to provide marginal coverage guarantees for the prediction sets on unseen data sampled from the global distribution, $\mathbb{Q}_{\text{test}} = \sum_{k=1}^{K}\lambda_k \mathbb{P}_{k}$, where $\lambda$ is a probability vector.
    
\begin{figure*}[!t]
\centering
\begin{subfigure}{\textwidth}
  \includegraphics[width=0.95\linewidth]{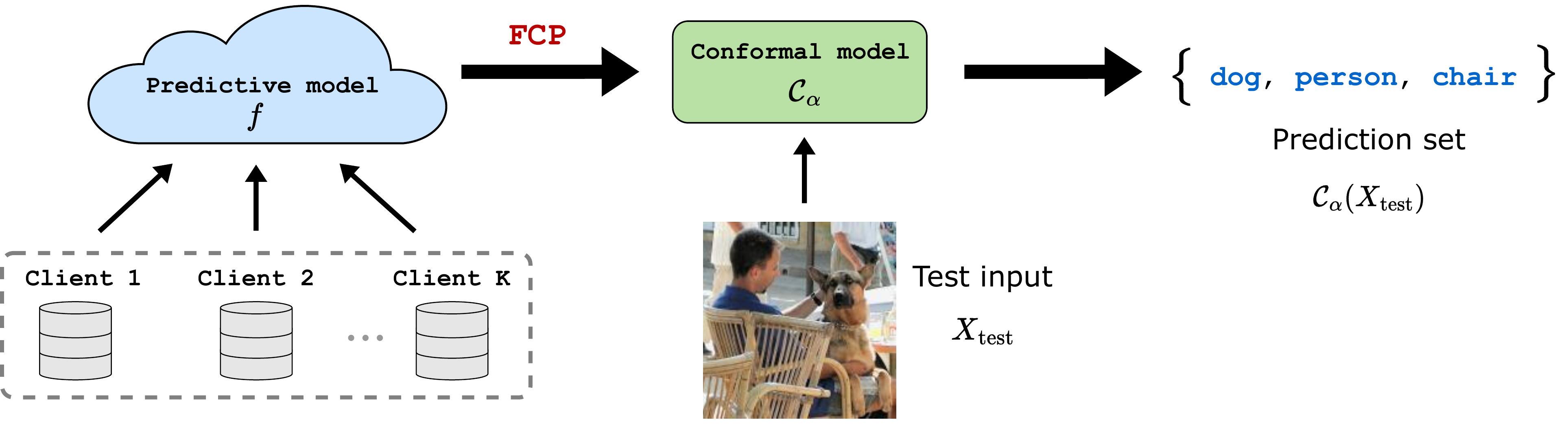}
\end{subfigure}
\vspace{-0.15in}
    \caption{\textbf{Overview of federated conformal prediction.} Given $K$ clients and a federated model $f$, we can obtain a federated conformal model $C_{\alpha}$ in the distributed environment. The conformal model $\mathcal{C}_{\alpha}$ produces a prediction set $\mathcal{C}_{\alpha}(X_{\text{test}})$ for an unseen test sample $(X_{\text{test}}, Y_{\text{test}})$. Prediction sets $\mathcal{C}_{\alpha}(X_{\text{test}})$ will contain the true label $Y_{\text{test}}$ with probability $1-\alpha$, i.e.,  $\mathbf{P}\left(Y_{\text{test}} \in \, \mathcal{C}_\alpha(X_{\text{test}})\right) \geq 1 - \alpha$. \textbf{\texttt{FCP}} represents our proposed federated conformal prediction method. Refer to Algorithm~\ref{alg:fcps} for details on learning federated conformal models.}
    \label{fig:fig1}
    \vskip -0.1in
\end{figure*}

However, we show that the natural heterogeneity in FL among the client distributions $\{\mathbb{P}_k\}$ raises numerous issues. First, heterogeneity immediately voids all standard conformal prediction guarantees since it contradicts exchangeability, a fundamental assumption in conformal prediction. Second, it also increases the size of the prediction sets, leading to less reliable or useful uncertainty estimates. Additional challenges in FL include potential uncertainty around the global distribution $\mathbb{Q}_{\text{test}}$~\cite{mohri2019agnostic} and requiring a fully distributed implementation with minimal communication~\cite{bonawitz2022federated}. We show how to carefully design novel methods to overcome these challenges and propose the following:
\begin{enumerate}
% \vspace{-0.1in}
    \item A framework for federated conformal prediction,
    \item A theoretical extension of conformal prediction under partially exchangeable client distributions, inexact quantile computations, and uncertain test distributions, 
    \item Thorough empirical evaluations and ablations under data heterogeneity on several benchmark computer vision and medical imaging datasets.
\end{enumerate}

\section{Preliminaries}
\label{sec:background}
\subsection{Conformal Prediction}
\label{sub:cp}
Distribution-free uncertainty quantification techniques such as conformal prediction have emerged as a general framework for providing black-box models such as deep learning with rigorous statistical guarantees. Conformal prediction has been applied to a wide range of applications such as image segmentation, few-show-learning, generative adversarial networks~\cite{angelopoulos2021gentle,bates-rcps,fisch2021fewshot,sankaranarayanan2022semantic}.

The general goal behind conformal prediction is to construct a set-valued predictor $\mathcal{C}: \mathcal{X} \rightarrow 2^\mathcal{Y}$ in such a way as to ensure \emph{coverage} --- controlling some risk function (e.g., zero-one loss) at a desired confidence level --- and have \emph{efficiency} --- reducing the size of the prediction set --- while maintaining coverage.

\textbf{Coverage.} Suppose we have a classifier $f: \mathcal{X} \rightarrow \Delta^J$ that outputs probabilities for $J$ classes and a desired error rate $\alpha \in (0, 1)$.
Then, a prediction set $\mathcal{C}(X) \subseteq 2^\mathcal{Y}$ could be constructed by including only those classes in which the probability score exceeds the threshold $\tau=1-\alpha$ to form the prediction set ${ \mathcal{C}_\alpha(X) = \{j \in \mathcal{Y}: [f(X)]_j \geq \tau\} }$.
If we assume these scores are perfectly calibrated, then we would expect the true class to be an element of the resulting prediction set with probability at least $1 - \alpha$: 
\begin{equation}
\label{eq:marginal-coverage}
% \small
\mathbf{P}\left(Y \; \in \; \mathcal{C}_\alpha \left(X\right) \right) \geq 1 - \alpha.
\vspace{-0.02in}
\end{equation}
This property is called \textit{marginal coverage} (meaning it holds on average), and predictors that enjoy the guarantee in Eq.~\eqref{eq:marginal-coverage} are valid conformal predictors.
We can empirically evaluate coverage of a conformal predictor at a specific $\alpha$ as the average number of times the ground truth label is one of the elements in the prediction set for each example in some unseen test data ${ D_\text{test} = \left\{(X_i, Y_i) \right\}_{i=1}^M }$:
\begin{equation}
\label{eq:metric-coverage}
% \small
\text{Coverage}(\mathcal{C}_\alpha) = \frac{1}{M} \sum_{i=1}^M \mathbb{1}\{Y_i \in \mathcal{C}_\alpha (X_i)\}.
\end{equation} 

\textbf{Conformal procedure.} However, deep learning models have been shown to be poorly calibrated~\cite{pmlr-v70-guo17a,ovadia2019can}, so an arbitrary model will not be a valid conformal predictor with a marginal coverage guarantee.
Therefore, to \textit{conformalize} a model into a valid conformal predictor, we can use the procedure of split conformal prediction to estimate a threshold $\hat{\tau}$ on a \emph{held-out calibration dataset} ${ D_\text{cal} = \left\{\left(X_i, Y_i\right)\right\}_{i=1}^n \sim \mathcal{X} \times \mathcal{Y} }$ that is assumed to be exchangeable with unseen test data $(X_\text{test}, Y_\text{test})$.

\textbf{Conformal score functions.} 
Assume we have some \textit{conformal score function} ${ S: \mathcal{X} \rightarrow \mathbb{R}^+ }$, where lower values indicate more ``conformity'' between the test point and the calibration points. 
One example of a score function is the \textit{least ambiguous set-value classifier} (LAC), defined as ${ S(X, Y) = 1 - [f(X)]_Y }$, where $[f(X)]_Y$ is the softmax score of the true class label.
Two other score functions, namely Adaptive Prediction Sets (APS) and Regularized Adaptive Prediction Sets (APS), are discussed in Appendix~\ref{app:conformal-score-functions}.
Then, $\hat{\tau}$ can be estimated by taking the $(1-\alpha)$-quantile of conformal scores on the calibration data.
The resulting prediction sets ${ \mathcal{C}_\alpha(X) = \left\{y \in \mathcal{Y}: S(X, Y) \leq \hat{\tau}\right\} }$ will satisfy Eq.~\eqref{eq:marginal-coverage} at the desired miscoverage level $\alpha$.
Note that marginal coverage does not imply the stronger statement of \textit{conditional coverage}:
\vspace{-0.05in}
\begin{equation}
\label{eq:conditional-coverage}
% \small
\mathbf{P}\left(Y \in \mathcal{C}(X) \mid X = x\right) \geq 1 - \alpha,
\end{equation}
which is not generally possible to guarantee without strong modeling assumptions~\cite{vovk2012conditional}.

\textbf{Efficiency.} To be of practical use, we would prefer a conformal predictor that achieves marginal coverage \textit{efficiently}
\footnote{Otherwise, a predictor could always output the entire space of labels $\mathcal{Y}$ for every input and trivially satisfy marginal coverage.}.
A valid predictor is said to be efficient if the expected size of its prediction sets $\mathbb{E}[\lvert C_\alpha\left(X_\text{test}  \right)\rvert]$ is small.
We evaluate the efficiency of a predictor by the average size of its prediction sets on the test set:
\begin{equation}
\label{eq:metric-size}
% \small
\text{Size}\left(\mathcal{C}_\alpha\right) = \frac{1}{M} \sum_{i=1}^M \big| \mathcal{C}_\alpha(X_{i})\big|.
\end{equation}

\begin{figure*}[!t]
% \vskip 0.2in
\begin{center}
\centering
\begin{subfigure}{.50\textwidth}
\centering
\includegraphics[width=\linewidth]{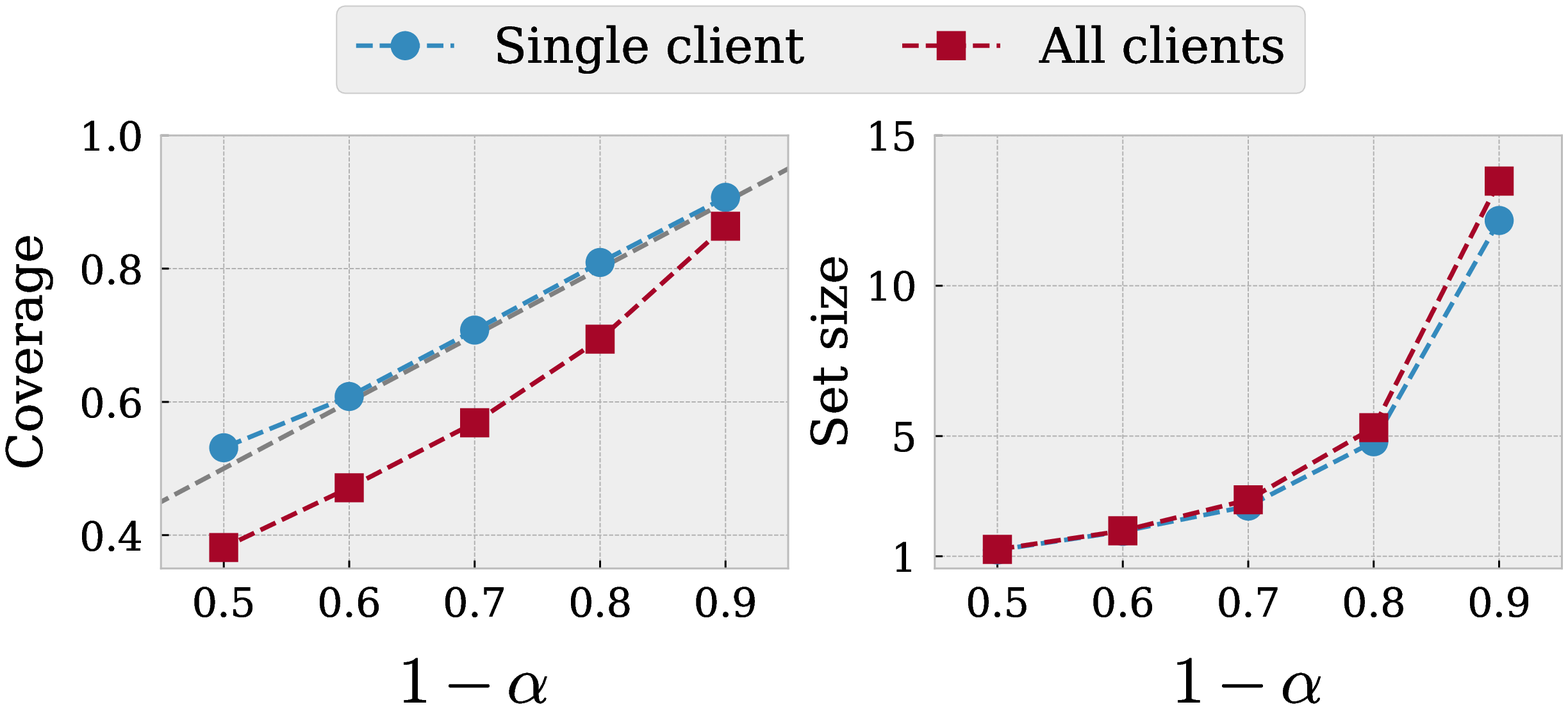}
\caption{\textbf{Testing for exchangeability --- local coverage is insufficient to guarantee global coverage.}}
\label{fig:challenge-exchangeability}
\end{subfigure}\hfill
\begin{subfigure}{.50\textwidth}
\centering
\includegraphics[width=0.9\linewidth]{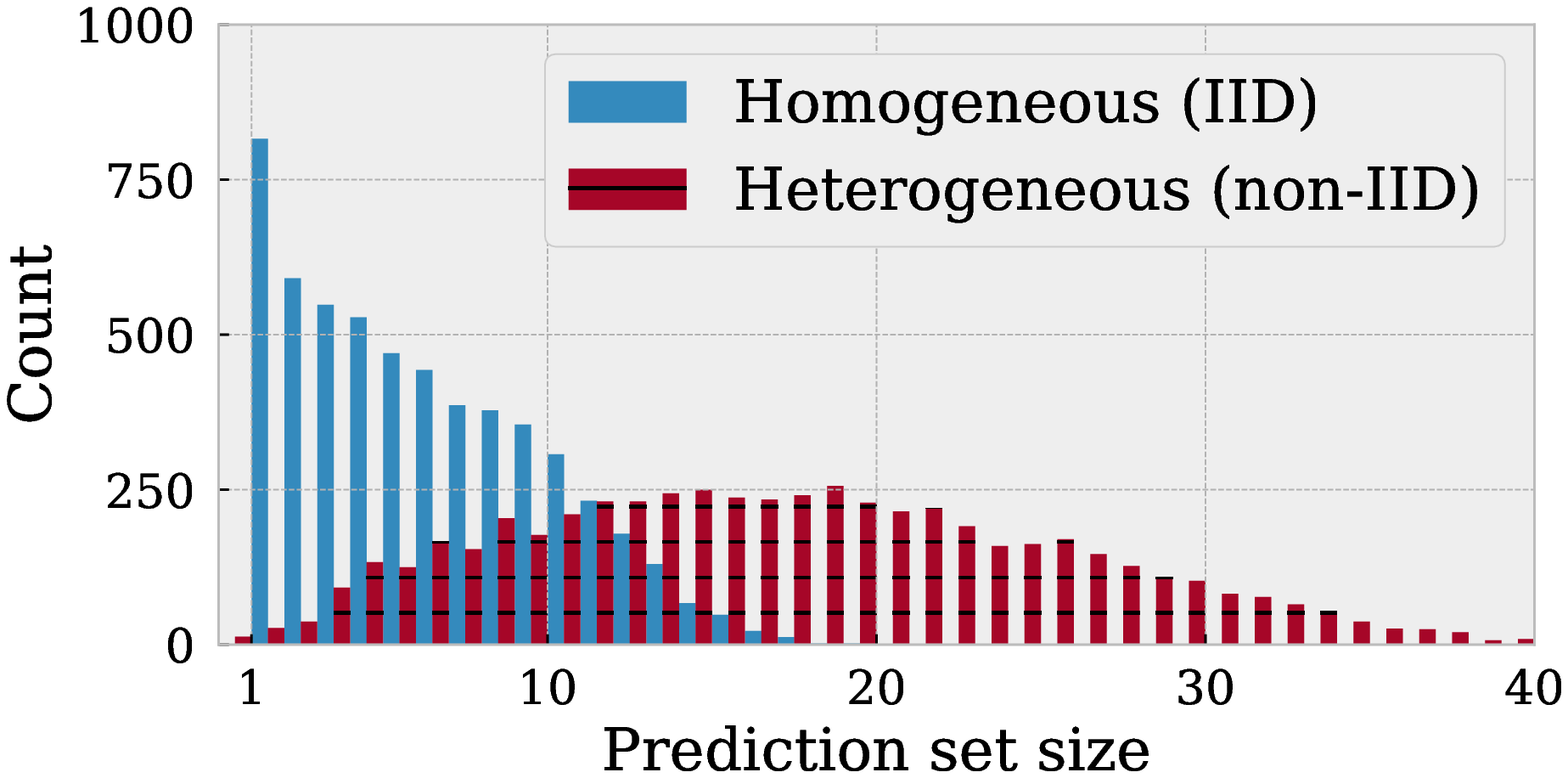}
\caption{\textbf{Non-IID clients degrade efficiency (increases set size).}}
\label{fig:challenge-efficiency}
\end{subfigure}
\vspace{-0.20in}
\caption{We trained ResNet-14 FedAvg models with 20 clients on CIFAR-100. 
(\subref{fig:challenge-exchangeability}) We trained a FedAvg model on CIFAR-100 with 20 non-IID clients and constructed predictors conformalized only on a single client, and we then evaluated the coverage on the first client's test set and all clients' test set. We see marginal coverage is satisfied only for the calibrated client and not for the aggregated clients. (\subref{fig:challenge-efficiency}) We trained two FedAvg models, one with IID clients and the other with non-IID clients. We plotted the distribution of predictions by set size and observed that predictors trained with heterogeneous clients have significantly lower efficiency than predictors trained with homogeneous clients.
% To demonstrate the effect of data heterogeneity on the efficiency of conformal prediction, 
}
\label{fig:challenges}
\end{center}
\vskip -0.15in
\end{figure*}

\subsection{Federated Learning} 
\label{sub:FL}
The success of deep learning can, in part, be attributed to its ability to accommodate increasingly large-scale datasets to improve predictive performance and unlock new capabilities not previously achievable at smaller scales~\cite{kaplan2020scaling,alabdulmohsin2022revisiting}.
However, in some domains, such as healthcare and finance, collecting and sharing large amounts of sensitive data may not be possible due to factors such as privacy concerns and financial incentives.
For this reason, federated learning (FL) is seen as a potential workaround for institutions in data-restricted domains to collaboratively develop models without transmitting sensitive or confidential data to an external party~\cite{Rieke_2020,terrail2022flamby}.

In FL, $K$ clients, each with their own individual datasets, attempt to optimize a global loss function ${L}$ that is the weighted average of local risk function $\ell_k$:
\begin{equation*}
    \underset{\theta}{\min}\;\Big\{ {L}(\theta) = \sum_{k=1}^K \lambda_k\cdot \mathbb{E}_{(x^k,y^k) \sim \mathbb{P}_k} [\,\ell_k(\theta; x^k, y^k)\,]\Big\},
\end{equation*}
where $P_k$ is the local data distribution on the $k$-th client, and $\lambda \in \Delta^K$ represents the weights. The most common choice of weights is $\lambda_k \propto O(n_k)$ (i.e. weighted by the empirical frequency), or uniform $\lambda_k = {1}/{K}$ \cite{mcmahan2017communicationefficient,wang2020tackling}.

Federated learning is most useful when clients have complementary datasets.
For example, data on patients with rare diseases may only be collected at a specialized treatment center, and data on the general patient population may be available at large hospitals.
With federated learning, these smaller centers and larger hospitals can collaboratively train a model to screen for the rare disease in the wider population. However, such \emph{data heterogeneity} across the clients is known to degrade prediction accuracy~\cite{zhao2018federated,hsieh2020non,karimireddy2020scaffold,li2022federated,hsu2019measuring,li2019convergence,zhu2021federated}. 
To address heterogeneity in federated learning, \citet{yu2022tct} recently proposed Train-Convexify-Train (TCT), which uses a two-stage approach that trains a FedAvg model in the first stage and then optimizes a convex approximation of the model based off the empirical (NTK) in the second stage.
They find the resulting model improves predictive accuracy significantly.
However, no prior work proposes to use conformal prediction with an FL model nor considers the effect of data heterogeneity on conformal predictors.

\section{Challenges of Federated Conformal Prediction}
\label{sec:effect}

Extending conformal prediction to the distributed setting is complicated by three main challenges: the lack of exchangeability between client distributions to ensure proper coverage guarantee, reduced efficiency of conformal predictors in highly data heterogeneous environments, and a communication-efficient distributed implementation of the conformal procedure. 

\subsection{Violation of Exchangeability}
The main assumption conformal prediction relies upon is exchangeability between the calibration data distribution and the test data distribution during inference.
Having exchangeable random variables implies identical (but not necessarily independent) distributions. 
Imagine the setting where $K$ clients have samples drawn IID from the same distribution $(X^k, Y^k) \sim \mathbb{P}, \forall k$. 
Then, the threshold computed on any one client could be used to conformalize every other client while ensuring valid coverage (e.g., the client with the largest amount of calibration samples calculates the $(1-\alpha)$  empirical quantile and shares this quantile with the server, which broadcasts it to the rest of the clients to construct conformal predictors).

However, this assumption of identical distributions is certain to be violated in real-world FL~\cite{terrail2022flamby}. 
For example, if clients have significant label skew or only have data from some subset of classes, then calibrating on a single client would not provide the correct coverage on other clients (see Figure~\ref{fig:challenge-exchangeability}). 
Instead, guaranteeing coverage on the global distribution would require tackling non-exchangeable client data distributions.

\subsection{Decreased Efficiency of Conformal Predictors}
A conformal predictor is efficient if it achieves marginal coverage with prediction sets with a small set size.
However, data heterogeneity can result in decreased efficiency of conformal predictors.
In Figure~\ref{fig:challenge-efficiency}, we trained two versions of FedAvg with 20 clients on CIFAR-100.
We assigned five classes to each client for the heterogeneous model, while for the homogeneous model, we randomly assigned classes to all clients.
We found the prediction sets of the heterogeneous model had an average set size of 7.5, while the prediction sets of the homogeneous model had a much higher average set size of 21.1.
For many practical applications, improving the efficiency of federated conformal predictors under data heterogeneity will be a crucial consideration.

\subsection{Distributed Implementation}
To ensure correct coverage for non-IID clients, the conformal score quantile must be estimated on calibration data from all clients in a distributed manner.
If all clients send their respective calibration scores, computing the quantile on the central server would be straightforward.
However, there are several reasons why a distributed quantile algorithm would be preferable to central aggregation.
Privacy concerns underlie the motivation of federated learning~\cite{bonawitz2022federated}, and sharing the score distribution from all the clients could potentially comprise  privacy. 
Instead, clients should strive to minimize all unnecessary communications.

\section{Methods}
\label{sec:methods}
In this section, we first introduce the exchangeability assumption in federated learning, which is slightly different from the assumption made in non-FL conformal prediction methods~\cite {vovk2005algorithmic, lei2017distributionfree}. Next, we propose the federated conformal prediction method with provable finite-sample coverage on unseen test samples. 
We then study how to improve the communication efficiency of the proposed method in the distributed environment by leveraging techniques on distributed quantile estimation~\cite {luo2016quantiles,masson2019ddsketch}. To this end, we present our algorithm in Algorithm~\ref{alg:fcps}.

\begin{algorithm}[tb]
\small
\caption{Federated Conformal Prediction (FCP)}
\label{alg:fcps}
\textbf{Input}: 
    global model $f_{\theta}$ parameterized by weights $\theta$, $R$ optimization rounds,
    $K$ training sets $\{(\hat{X}_i^k, \hat{Y}_i^k)\}_{i=1}^{m_k},\; k \in [K]$, 
    $K$ validation sets $\{(X_i^k, Y_i^k)\}_{i=1}^{n_k},\; k \in [K]$, 
    conformal score function $S: \Delta^J \rightarrow \mathbb{R}^+$, error threshold $\alpha$, federated optimization algorithm  \textsf{FedOpt} (we recommend applying TCT~\citep{yu2022tct} with logistic regression).
\\
\textbf{Output}: Set-valued function $\mathcal{C}_\alpha(\cdot)$ \\
\vspace{-0.15in}
\begin{algorithmic}[1]
    \STATE \texttt{// Step-1: learn predictive FL model}
    \vspace{0.05in}
    \STATE $f \leftarrow \textsf{FedOpt}(f_{\theta}, \{(\hat{X}_i^k, \hat{Y}_i^k)\}_{i\in[m_k], k\in[K]}, R)$
    \vspace{0.05in}
    \STATE \texttt{// Step-2: construct federated conformal predictor}
    \FOR{$k \in \{1, 2, \ldots, K\}$}
        \FOR{$i \in \{1, 2, \ldots, n_k\}$}
            \STATE \texttt{// Compute conformal score}
            \STATE $s_i = S(f(X_i), Y_i))$
        \ENDFOR 
        \STATE \texttt{// Sketch and communicate scores}
        \STATE $\hat{s}_k \leftarrow\textsf{Sketch}(\{s_i\}_{i=1}^{n_k})$ 
        \STATE Communicate sketch of scores $\hat{s}_k$ to central server 
    \ENDFOR 
    % \STATE Merge the $K$ sketches
    \STATE \texttt{// distributed quantile estimation}
    \STATE $\hat{q}_\alpha := \textsf{DistributedQuantile}\left(\{\hat{s}_k\}_{k=1}^{K}, \frac{\lceil(1-\alpha)(N + K)\rceil}{N}\right)$  
    % \STATE $\hat{q}_\alpha := \left(\frac{\lceil(1-\alpha)(N + K)\rceil}{N}\right)$  \hfill\COMMENT{// Compute quantile on merged sketches}
    \STATE For $X\in \mathcal{X}$, ${ \mathcal{C}_{\alpha}(X) := \left\{y \in \mathcal{Y}: S(X, y) \leq \hat{q}_{\alpha}\right\} }$
    \STATE \textbf{Return} $\mathcal{C}_{\alpha}(\cdot)$
\end{algorithmic}
\end{algorithm}

\subsection{Conformal Prediction with Partial Exchangeability}\label{subsec:fl-conformal}
Recall that we denote the distribution of the $k$-th client by $\mathbb{P}_{k}$, i.e., $(X_i^{k}, Y_i^{k}) \sim \mathbb{P}_{k}$, where $\{(X_i^{k}, Y_i^{k})\}_{i=1}^{n_k}$ are the $n_k$ held out calibration samples from the $k$-th client. Suppose the future test point $(X_\text{test}, Y_\text{test})$ is sampled from the global distribution $\mathbb{Q}_{\text{test}} = \mathbb{Q}_{\lambda}$ for some probability vector $\lambda \in \Delta^K$,
\begin{equation}\label{eq:lambda-test-distribution}
    (X_\text{test}, Y_\text{test}) \stackrel{\text{i.i.d.}}{\sim} \mathbb{Q}_{\lambda},\quad \mathbb{Q}_{\lambda} = \sum_{k=1}^{K}\lambda_k \mathbb{P}_{k}.
\end{equation}
This implies that the global distribution $\mathbb{Q}_{\lambda}$ is drawn from $\mathbb{P}_{k}$ with probability $\lambda_k$. Hence, with probability $\lambda_k$, it has the same distribution as the data points on client $k$. This forms the basis of our assumption, stated informally below. A more formal version is detailed in Appendix~\ref{subsec:app-partial-exchange}.

\begin{assumption}[Exchangeability in FL]\label{assumption:fl-exchangeable}
For a probability vector $\lambda \in \Delta^K$, the scores on client $k$: ${ S(X_{1}^{k}, Y_{1}^{k}), \dots, S(X_{n_k}^{k}, Y_{n_k}^{k}), S(X_\text{test}, Y_\text{test}) }$ are exchangeable with probability $\lambda_k$.
\end{assumption}
\begin{remark}
The above assumption can be interpreted as a variant of \textit{partial exchangeability}~\citep{de1980condition, diaconis1988recent}, which is a generalization of exchangeability.
It makes no assumptions between $\mathbb{P}_1, \dots, \mathbb{P}_K$. Specifically, this assumption does not require independence or identical distributions among the clients.
\end{remark}

As an example of how Assumption~\ref{assumption:fl-exchangeable} may be better suited than the standard exchangeability assumption, imagine several different hospitals participating in an FL collaboration, where each hospital is a client.
Clearly, each hospital will have a different underlying patient population and different data acquisition processes so that each client will have a different data distribution but patients from the same hospital will be assumed to be exchangeable.
Also, independence may be violated between clients as some subset of patients can be treated at multiple hospitals.

\begin{theorem}\label{theorem:fl-conformal}
Under Assumption~\ref{assumption:fl-exchangeable}, suppose there are $n_k$ samples from the $k$-th client, $N=\sum_{k=1}^{K}n_k$, and $\lambda_k \propto (n_k + 1)$. Define $\hat{q}_{\alpha}$ as the ${\lceil (1-\alpha) (N+K) \rceil}$ largest value in $\{(S(X_{i}^{k}, Y_{i}^{k}))\}_{i\in [n_k],\, k\in[K]}$ and \begin{equation*}
\mathcal{C}_{\alpha}(X) = \left\{y \in \mathcal{Y}: S(X, y) \leq \hat{q}_{\alpha}\right\}.
\end{equation*}
Then, $C_\alpha(\cdot)$ is a valid conformal predictor with
\begin{equation}\label{eq:theorem-results}
     1 - \alpha + \frac{K}{N+K} \geq \mathbf{P}\left(Y_\text{test} \in \, \mathcal{C}_\alpha(X_\text{test})\right) \geq 1 - \alpha \,.
\end{equation}
\end{theorem}
The proof of Theorem~\ref{theorem:fl-conformal} can be found in Appendix~\ref{subsec:proof-of-main-theorem}. 
As shown in Eq.~\eqref{eq:theorem-results}, our proposed method is guaranteed to achieve $(1-\alpha)$ marginal coverage. 

\textbf{Comparison with the IID Setting.} Observe that the gap between the upper and lower bound in Eq.~\eqref{eq:theorem-results} is $\frac{K}{N+K} = \frac{1}{N/K + 1}$. 
Thus, the gap depends on the \emph{average} number of data points per client. If a certain client $k$ has very few data points with a small $n_k$, we will have high uncertainty about the predictions corresponding to client $k$. This is compensated by client $k$ having a lower weight in our test distribution since $\lambda_k \propto (n_k+1)$. As a result, our approach is suitable for settings where the average number of data points per client is large. In particular, for non-vacuous bounds, we need
\begin{equation*}
     \lceil(1 - \alpha)(N+K)\rceil \leq N \Rightarrow \alpha \geq \frac{1}{N/K + 1}\,.
\end{equation*}
If, instead, we had IID data points, with all data points being exchangeable, \citet{lei2018distribution} show that we can compute quantiles with a guarantee of
\[
(1-\alpha) \leq \mathbf{P}\left(Y_\text{test} \in \, \mathcal{C}_\alpha(X_\text{test}) \right) \leq\, (1-\alpha) + \frac{1}{N+1}\,.
\]
This results in a gap of $\frac{1}{N+1}$. Our result recovers this as a special case with $K=1$ but degrades with increasing $K$. This looseness in the analysis translates to needing a larger quantile. In the IID setting with full exchangeability assumption, we need to pick $\hat{q}_{\alpha}$ to be the ${\lceil (1-\alpha) (N+1) \rceil}$ largest value. In contrast, with partial exchangeability, we need the ${\lceil (1-\alpha) (N+K) \rceil}$ largest value.

\textbf{Discussion about $\lambda$.} As described in Theorem~\ref{theorem:fl-conformal}, we assume that $\lambda_k \propto (n_k + 1)$, where $\lambda$ is a known probability vector for defining the test distribution $\mathbb{Q}_{\lambda}$. 
In the federated learning literature, the test distribution is normally defined as the mixture of the distributions of different clients, and the weight of each distribution is proportional to the number of samples from that client~\citep{mcmahan2017communicationefficient,wang2020tackling}. 
On the other hand, our method can be extended to the setting where $\lambda$ is unknown during test time~\citep{mohri2019agnostic}. 
Suppose $\lambda$ is constrained in a convex set $\Lambda$, for example, $\Lambda = \{\hat{\lambda} | \hat{\lambda}_k \geq (1-\delta)(n_k+1)/(N+K), k \in [K]\}$, we can modify the definition of $\hat{q}_{\alpha}$ in our algorithm to achieve valid converge even when $\lambda$ violates the assumption $\lambda_k \propto (n_k + 1)$. 
The modified federated conformal method and the theoretical results  can be  found in Appendix~\ref{subsec:appendix-extension-to-unkown-lambda}.

\subsection{Distributed Quantile Estimation}\label{sub:quantile}
To learn the set-valued function $\mathcal{C}_{\alpha}$ in our  method, we need to compute the quantile of the conformal scores $\{s_{i}^{k}\}_{i\in [n_k],\, k\in[K]}$ that are distributed across $K$ clients, where $s_i^{k} = S(X_i^{k}, Y_i^k)$.
In order to reduce the number of communications for estimating the empirical quantiles in our method, we apply distributed quantile estimation techniques~\cite {luo2016quantiles, DUNNING2021100049}.

In particular, we utilize T-Digest, a quantile sketching algorithm well suited for computing online quantile estimates for distributed workflows~\cite{DUNNING2021100049}.
Importantly, this data structure is mergeable, which allows for distributed aggregation in parallel workflows.

As the quantile is approximately computed in the distributed quantile estimation methods, in what follows, we provide the coverage guarantees of our method when using inexact quantiles. 
We first introduce the $\varepsilon$-approximate $\beta$-quantile~\citep{luo2016quantiles}.
\begin{definition}[$\varepsilon$-approximate $\beta$-quantile]\label{def:approx-quantile}
For an error $\varepsilon \in (0, 1)$, the $\varepsilon$-approximate $\beta$-quantile is any element with rank between $(\beta-\varepsilon)N$ and $(\beta+\varepsilon)N$, where $N$ is the total number of elements.
\end{definition}
If the distributed quantile estimation method outputs $\varepsilon$-approximate $\beta$-quantiles, our method approximately achieves desired coverage.
\begin{corollary}\label{corollary:fl-conformal-approx}
Under Assumption~\ref{assumption:fl-exchangeable}, suppose there are $n_k$ samples from the $k$-th client, $N=\sum_{k=1}^{K}n_k$, and $\lambda_k \propto (n_k + 1)$, and the sketch algorithm outputs the $\varepsilon$-approximate $(1-\alpha)$-quantile. Then the output $\mathcal{C}_{\alpha}$ of Algorithm~\ref{alg:fcps} satisfies
\begin{equation}\label{eq:corollary-distributed-quantile}
\begin{aligned}
\mathbf{P}\left(Y \in \, \mathcal{C}_\alpha(X)\right) &\geq 
1 - \alpha - \hat{\varepsilon} - \frac{\mathbb{1}\{\varepsilon > 0\}}{N+K},\\
\mathbf{P}\left(Y \in \, \mathcal{C}_\alpha(X)\right) &\leq 
1 - \alpha + \hat{\varepsilon} + \frac{K}{N+K},
\end{aligned}
\end{equation}
where $\hat{\varepsilon} = {\varepsilon N}/{(N+K)}$.
\end{corollary}
As suggested by the above result, when $\varepsilon$ is small, Algorithm~\ref{alg:fcps} achieves similar coverage compared to Theorem~\ref{theorem:fl-conformal} with the exact quantile. When the approximation error $\varepsilon=0$, the results in Corollary~\ref{corollary:fl-conformal-approx} reduces to the ones in Theorem~\ref{theorem:fl-conformal}. To further elucidate the above theoretical results, we take the algorithm from \citep{huang2011sampling} as an example and provide a detailed theoretical statement in Appendix~\ref{sec:appendix-dist-quantile}.

We also conduct experiments to study the performance of different quantile estimation methods in our algorithm. The results are summarized in Figure~\ref{fig:quantile-compare}, and we find that T-digest performs similarly to the exact quantile.  
Specifically, for T-digest, the accuracy of estimating the $q$-quantile is approximately constant relative to $q\cdot(1-q)$, and memory of $\Theta(\delta)$, where $\delta$ is a compression parameter $\delta \ll n$.
This relative error bound is well-suited for computing quantiles near $0$ or $1$, which is often necessary with small $\alpha$ values, compared to absolute error bounds of other quantile approximation methods such as DDSketch~\citep{masson2019ddsketch}.  
In other words,  T-digest achieves smaller $\hat{\varepsilon}$ in Eq.~\eqref{eq:corollary-distributed-quantile} when $\alpha$ is close to 1.

\begin{figure*}[ht!]
% \vskip 0.2in
\begin{center}
  \centering
\begin{subfigure}{.5\textwidth}
  \centering
  \includegraphics[width=\linewidth]{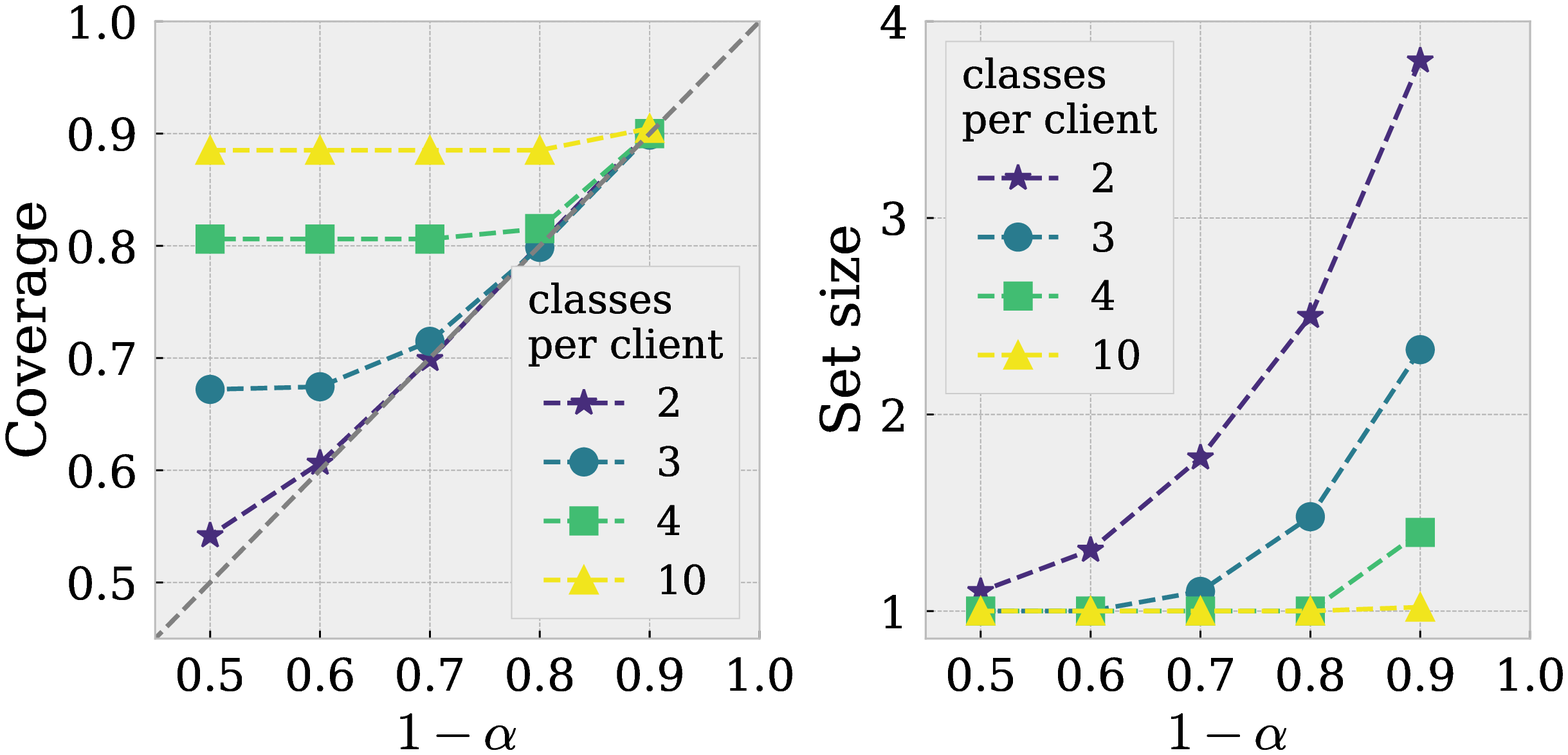}
  \caption{\textbf{Federated Conformal Predictor using FedAvg.}}
  \label{fig:fedavg}
\end{subfigure}\hfill
\begin{subfigure}{.5\textwidth}
  \centering
  \includegraphics[width=\linewidth]{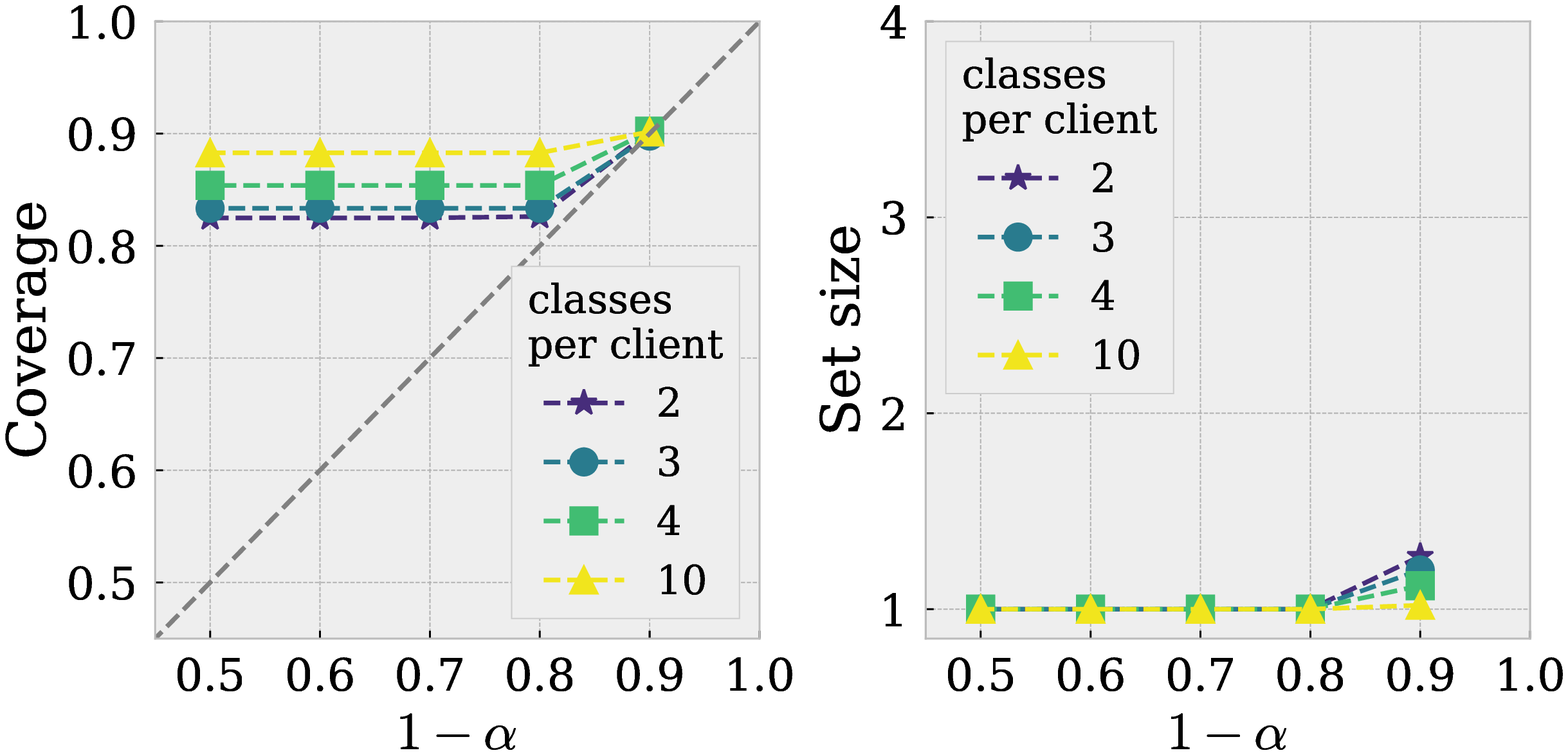}
  \caption{\textbf{Federated Conformal Predictor using TCT.}}
  \label{fig:tct-vs-fedavg}
\end{subfigure}
\caption{\textbf{Efficiency of conformal predictors with TCT is more robust under data heterogeneity.} We trained FedAvg and TCT with five clients on CIFAR-10 with four different amounts of data heterogeneity (two, three, four, or all ten classes) for each client. As heterogeneity increases, the average prediction set size also drastically increases for FedAvg but not for TCT.}
\end{center}
\vskip -0.1in
\end{figure*}

\section{Experiments}
\label{sec:experimets}
Our experiments evaluate the proposed federated conformal prediction (FCP) framework under different types of data heterogeneity.
We demonstrate an application of FCP for selective classification on a realistic skin lesion classification task.
We also perform a number of ablation experiments to evaluate the choice of conformal score function, quantile estimation method, and FL optimization.

\subsection{Datasets}
Our experiments use several computer vision datasets (FashionMNIST, CIFAR-10, and CIFAR-100) and medical imaging datasets (DermaMNIST, PathMNIST, TissueMNIST, and Fitzpatrick17K).

The Fitzpatrick17K skin lesion dataset is a challenging and real-world dataset with 114 different skin conditions with a top-1 accuracy of 20.3\% reported in the original paper~\cite{Groh2021EvaluatingDN}.
The 114 skin conditions are grouped into three broad disease categories: ``non-neoplastic'', ``malignant'', and ``benign''.
Additionally, each image is labeled with its Fitzpatrick skin type, rated on an ordinal six-point scale that measures skin types (lighter skin types have higher rates of skin cancer; see Figure~\ref{fig:skin-rate}). 
For this dataset, we experiment with two types of client heterogeneity: disease categories and skin types.

The clients for the rest of the datasets were partitioned by class.
For FashionMNIST~\cite{xiao2017fashion}, one class was assigned to each of the ten clients.
For CIFAR10~\citep{krizhevsky2009learning}, two classes were assigned to each of the five clients.
For CIFAR100, five classes were assigned to each of the 20 clients.
For MedMNIST~\cite{yang2023medmnist} datasets (DermaMNIST, PathMNIST, TissueMNIST) were partitioned by groups of 2 and 3 classes; see Appendix~\ref{app:data} for the exact partition.

\subsection{Experimental Setup}
\label{sub:setup}
For each dataset, we trained a centralized model and three federated models: FedAvg~\cite{McMahanMRA16}, FedProx~\cite{li2020federated}, and TCT~\cite{yu2022tct}.
For the decentralized models, we introduced heterogeneity by partitioning data based on the class between clients (skin type and disease category for the Fitzpatrick17K dataset).

It is standard practice to apply some form of scaling to calibrate the output of classifiers better, so we applied temperature scaling~\citep{guo2017calibration} to the logits where the temperature $T$ was learned by minimizing negative log-likelihood on the calibration set $D_\text{val} = \left\{\left(x_i, y_i\right)\right\}_{i=1}^N$,
\begin{equation*}
    \underset{T}{\min}\; L(T) = -\sum_{i=1}^N \sum_{j=1}^J \mathbb{1}_{\{y_i = j\}} \log\Big(\left[\sigma_{\text{s}}\left(f(x_i)/T\right)\right]_j\Big),
\end{equation*}
where $\sigma_{\text{s}}(\cdot)$ denotes the {softmax} function.  
For the decentralized models, we use the unweighted average of client temperatures ${ T = \frac{1}{K} \sum_{k=1}^K T_k }$.

We evaluated three nonconformity scores (LAC, APS, and RAPS) defined in Appendix~\ref{app:conformal-score-functions}.
For all score functions, we forced each prediction set to contain at least one prediction by always including the class with the highest score in the prediction set\footnote{As a consequence, coverage will never fall below top-1 classification accuracy, and empirical coverage may exceed the marginal coverage guarantee. The upper bound of marginal coverage holds if empty prediction sets are allowed.}. 
For each experiment, we report metrics over ten trials, where the calibration and test sets are randomly split evenly in each trial.

We provide implementation code \url{https://github.com/clu5/federated-conformal}.

\subsection{Main Results}
\label{sub:results}

Our results show that federated conformal predictors using TCT are more robust to data heterogeneity than other federated models such as FedAvg and FedProx.
We also demonstrate conformal prediction for the task of selective classification on the Fitzpatrick17k skin lesion dataset.

\begin{table}[t!]
    \vspace{-0.15in}
    \caption{\textbf{Relative inefficiency of decentralized conformal predictors over centralized baseline.} We compare the inefficiency, measured by the ratio of average prediction set size over the centralized baseline, of different decentralized methods for conformal prediction with LAC. Lower inefficiency is better (``1x'' would indicate a method is just as efficient as the centralized baseline); bold denotes the most efficient method.}
    \label{tab:relative-inefficiency}
    \begin{center}
    \begin{small}
    \begin{tabular}{lc|c|c|c}
    \toprule
    Dataset & $1 - \alpha$ & FedAvg & FedProx & TCT \\
    \midrule
    \multirow{2}{*}{FashionMNIST}
    & $0.90$ & 8.6$\times$& 9.3$\times$& \textbf{1.1$\times$} \\
    & $0.80$ & 8.0$\times$& 7.2$\times$& \textbf{1.0$\times$} \\
    \midrule
    \multirow{2}{*}{CIFAR-10}
    & $0.90$ & 3.5$\times$& 3.2$\times$& \textbf{1.2$\times$} \\
    & $0.80$ & 2.5$\times$& 2.3$\times$& \textbf{1.0$\times$} \\
    \midrule
    \multirow{2}{*}{CIFAR-100}
    & $0.90$ & 3.8$\times$& 3.8$\times$& \textbf{1.2$\times$} \\
    & $0.80$ & 4.1$\times$& 4.2$\times$& \textbf{1.2$\times$} \\
    \midrule
    \multirow{2}{*}{DermaMNIST}
    & $0.90$ & 3.1$\times$& 3.2$\times$& \textbf{2.4$\times$} \\
    & $0.80$ & 2.1$\times$& 2.2$\times$& \textbf{1.3$\times$} \\
    \midrule
    \multirow{2}{*}{PathMNIST}
    & $0.90$ & 3.2$\times$& 2.8$\times$& \textbf{1.2$\times$} \\
    & $0.80$ & 2.5$\times$& 2.0$\times$& \textbf{1.0$\times$} \\
    \midrule
    \multirow{2}{*}{TissueMNIST}
    & $0.90$ & 1.8$\times$& 1.8$\times$& \textbf{0.9$\times$} \\
    & $0.80$ & 2.0$\times$& 2.0$\times$& \textbf{0.9$\times$} \\
    \midrule
    \multirow{2}{*}{\shortstack[l]{Fitzpatrick17k\\ (skin type)}}
    & $0.90$ & 1.3$\times$& 1.3$\times$& \textbf{1.2$\times$} \\
    & $0.80$ & 1.5$\times$& 1.4$\times$& \textbf{1.2$\times$} \\
    \midrule
    \multirow{2}{*}{\shortstack[l]{Fitzpatrick17k\\ (disease category)}}
    & $0.90$ & 2.3$\times$& 2.2$\times$& \textbf{1.3$\times$} \\
    & $0.80$ & 3.1$\times$& 2.7$\times$& \textbf{1.3$\times$} \\
    \bottomrule
    \end{tabular}
    % }
    % \end{sc}
    \end{small}
    \end{center}
    \vskip -0.15in
\end{table}

\begin{figure}[t!]
    \begin{center}
    \centerline{\includegraphics[width=1.00\columnwidth]{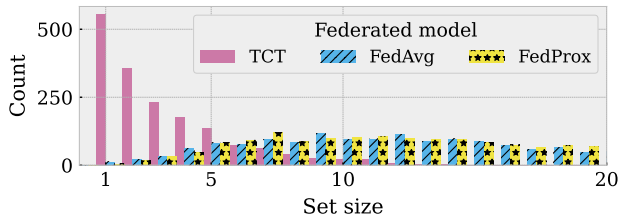}}
    \vspace{-0.1in}
    \caption{\textbf{TCT is more efficient than FedAvg and FedProx even on the same subset of predictions correctly classified by all methods.} We considered only the subset of CIFAR-100 that was correctly predicted by all three federated models (according to top-1 accuracy) and plotted the distribution of set sizes of their respective prediction sets with LAC at $\alpha=0.1$ (plot clipped past size 20).}
    \label{fig:compare-set-size-same-subset}
    \end{center}
    \vskip -0.4in
\end{figure}

\textbf{Comparing efficiency under heterogeneity.} 
In Figure~\ref{fig:tct-vs-fedavg}, we compared federated conformal predictors' coverage and set size using FedAvg and TCT under varying amounts of data heterogeneity.
As heterogeneity increases, prediction sets with the FedAvg model have larger set sizes, while prediction sets with the TCT maintain small set sizes.
In Table~\ref{tab:relative-inefficiency}, we measured the relative increase in average set size over centralized conformal predictors. We found that TCT has better efficiency than FedAvg and FedProx across all datasets.
More detailed results can be found in Table~\ref{tab:full-results} in Appendix~\ref{app:experiments}.

To better control for differences in predictive performance, we plotted the set size distribution of test examples that were classified correctly by all three decentralized models in Figure~\ref{fig:compare-set-size-same-subset}.
This shows that the better efficiency of TCT is not only a result of greater prediction accuracy but also due to being more calibrated. 
This advantage in calibration is also robust across different nonconformity score functions (see Table~\ref{tab:cifar-efficiency-score-compare} in Appendix~\ref{app:experiments}).

\begin{figure}[t!]
    \begin{center}
    \centerline{\includegraphics[width=1.00\columnwidth]{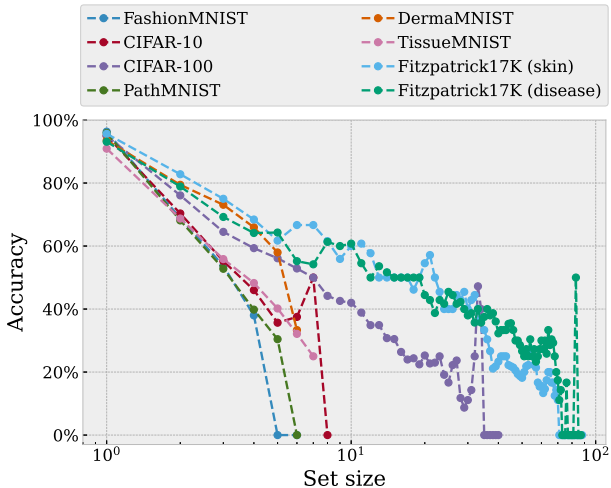}}
    \vspace{-0.05in}\caption{\textbf{Set size is negatively correlated with prediction accuracy}. We plotted the median of mean top-1 accuracy for each set size over 100 random trials for each dataset.}
    \label{fig:accuracy-vs-set-size}
    \end{center}
    \vskip -0.3in
\end{figure}
    
\textbf{Prediction set size correlates with accuracy.} 
    In Figure~\ref{fig:accuracy-vs-set-size}, we found a strong negative correlation between prediction set size and prediction accuracy for all datasets.
    This is intuitive as smaller sizes correspond to more confident predictions and lower conformality scores, while larger sizes correspond to less confident predictions and higher conformality scores.

\begin{figure}[t!]
    \begin{center}
    \centerline{\includegraphics[width=0.90\columnwidth]{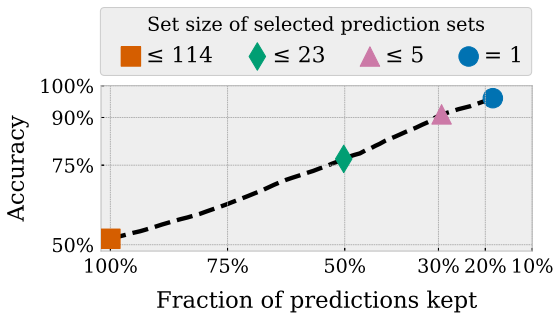}}
    \vspace{-0.05in}\caption{\textbf{Selective classification with conformal prediction.} We calibrate a TCT conformal predictor with RAPS score function at $\alpha-0.1$ on the Fitzpatrick17k skin lesion dataset. We plot the top-1 accuracy after the most uncertain predictions  (quantified by prediction set size) are excluded. From a baseline top-1 accuracy of 53\%, we can achieve 77\% accuracy when filtering out the 50\% largest sets, 90\% accuracy when filtering out the 70\% largest sets, and 95\% accuracy filtering out the 80\% largest sets.}
    \label{fig:selective-classification}
    \end{center}
    \vskip -0.3in
\end{figure}
    
\textbf{Selective classification with conformal.} 
One simple but useful application of conformal prediction is selective classification.
In some applications, such as cancer screening, we may prefer to make fewer, higher-quality predictions instead of outputting a prediction for every data point. 
In Figure~\ref{fig:selective-classification}, we plotted the increase in top-1 accuracy that can be achieved by excluding the predictions with a large set size.
While this approach is only a heuristic and does not have a coverage guarantee, prediction sets of small size empirically have high coverage (Table~\ref{tab:size-conditional-coverage}), conformal prediction can be extended to provide formal guarantees for controlling different types of risks such as false negative rate~\cite{angelopoulos2021learn}.

\subsection{Ablation Studies}
\label{sub:ablation}
We conducted several ablation experiments on federated conformal predictors and empirically found RAPS, T-Digest, and TCT optimized with cross-entropy to be good defaults for conformal score function, quantile sketcher, and federated optimization procedure, respectively.

\begin{figure}[t!]
    \begin{center}
    \centerline{\includegraphics[width=0.97\columnwidth]{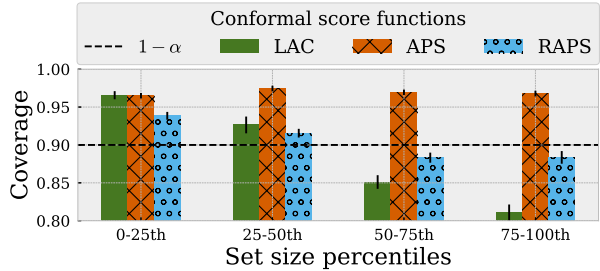}}
    \caption{\textbf{RAPS score function maintains tighter coverage at different set sizes than LAC and APS.} We evaluated the size conditional coverage and average set size of each quartile of set sizes and see that LAC has lower than $1-\alpha$ coverage on prediction with larger sizes. In comparison, RAPS has tighter $1-\alpha$ coverage across different set sizes.}
    \label{fig:conformal-score-compare}
    \end{center}
    \vskip -0.3in
\end{figure}

\textbf{Comparing score functions.} 
We can approximately evaluate conditional coverage as size-stratified coverage (SSC) as proposed by~\citet{angelopoulos2020sets}:
In some sense, this measures a predictor's adaptiveness to different inputs, meaning that larger sets represent more difficult or uncertain predictions and smaller sets represent easier or more confident predictions~\cite{angelopoulos2021gentle}.
We fixed TCT as our model and evaluated different choices of the conformal score function.
We observed that RAPS has tighter coverage when stratified by set size, shown in Table~\ref{tab:size-conditional-coverage}, while LAC has too much coverage on small sets and too little coverage on large sets.

\begin{figure}[t!]
    \begin{center}
    \centerline{\includegraphics[width=1.00\columnwidth]{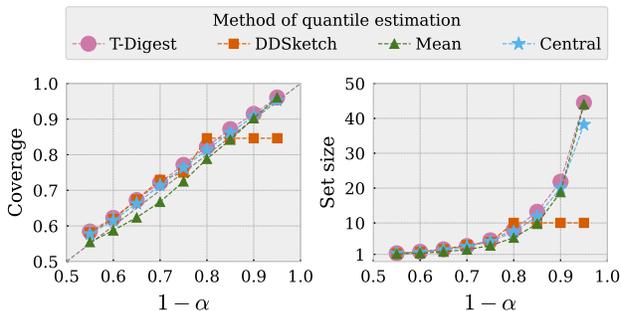}}
    \vspace{-0.1in}
    \caption{\textbf{Comparing different methods of distributed quantile approximation on Fitzpatrick17k.} Naively averaging client quantiles produces a biased estimate that does not provide correct coverage at lower thresholds (Mean). DDSketch has large errors computing quantiles at high $\alpha$ values. Only T-Digest closely approximates the true quantile of the centralized baseline.}
    \label{fig:quantile-compare}
    \end{center}
    \vskip -0.25in
\end{figure}

\newpage

\textbf{Comparing quantile methods.} 
In Figure~\ref{fig:quantile-compare}, we compared four different methods of computing approximate quantiles on Fitzpatrick17k with TCT and LAC.
Simply averaging client quantiles produces a biased estimate which does not provide correct coverage, while DDSketch, another distributed quantile estimator, results in bad quantile estimates at high confidence thresholds.
Only T-Digest produces the expected coverage of the centralized quantile across $\alpha$ thresholds.
    
\begin{table}[t]
\small
\caption{\textbf{Optimizing TCT with squared loss results in inefficient conformal predictors.} Comparing the average set size of TCT optimized with different stage 2 loss functions on CIFAR-100 at ${ \alpha = 0.1 }$ across conformal score functions (LAC, APS, and RAPS).}
\label{tab:tct-compare-loss}
\vskip 0.15in
\begin{center}
\begin{small}
\begin{sc}
\begin{tabular}{c|cccc}
\toprule
\shortstack[c]{loss function} & \shortstack[c]{Accuracy} & LAC & APS & RAPS\\
\midrule
Cross-entropy & $61\%$ & $6.5$ & $18.0$ & $8.3$ \\
\midrule
Squared loss & $58\%$ & $12.1$ & $78.8$ & $22.5$ \\
\bottomrule
\end{tabular}
\end{sc}
\end{small}
\end{center}
\vskip -0.2in
\end{table}

\textbf{Impact of FL optimization.} 
In Table~\ref{tab:tct-compare-loss}, we compared squared loss and cross-entropy loss for TCT optimization. We found that while the squared loss model had higher top-1 classification accuracy, the resulting prediction sets were much larger compared to models optimized with cross-entropy. 
We further investigated the difference in calibration error between losses in Figure~\ref{fig:loss-comparison-cifar100} and found that models optimized with squared loss result in extremely small softmax values, which have poor calibration even after temperature scaling.

\section{Conclusion and Future Work}\label{sec:conclude}

Conformal prediction is especially well suited to endow federated learning models with finite-sample coverage guarantees that can be used for downstream tasks, e.g., selective classification.
This paper introduced conformal prediction to the distributed learning setting with non-IID clients. We extended the statistical guarantees of marginal coverage to the mixtures of client distributions in federated learning. We also proposed efficient distributed algorithms to compute these conformal predictors.
We extensively evaluated the proposed federated conformal predictors under various data heterogeneity conditions over several computer vision benchmark datasets and medical imaging datasets.

In this work, we assumed that the future test distribution is $\mathbb{Q}_{\text{test}}$ is a mixture of the training distribution, but we do not discuss where the mixture weights come from. In practice, we would fit a mixture model using an empirical Bayes approach~\cite{mcauliffe2006nonparametric}. The weights can also be chosen adaptively based on the test data point~\cite{yu2022robust}. Further, the FCP framework can be extended straightforwardly to a hierarchical model where all the clients are drawn uniformly from the same meta-distribution. 
Finally, conformal prediction with personalized federated models is an important and challenging direction we are currently exploring.

\section*{Acknowledgements}
Charles Lu and Ramesh Raskar were partially supported by the MIT Media Lab Consortium and the National Science Foundation (NSF). Yaodong Yu, Sai Praneeth Karimireddy, and Michael I. Jordan were partially supported by the European Research Council (ERC) Synergy Grant program and the Mathematical Data Science program of the Office of Naval Research. Sai Praneeth Karimireddy is additionally supported by a Swiss National Science Foundation (SNSF) Fellowship.

\bibliography{icml}
\bibliographystyle{icml2023}

%%%%%%%%%%%%%%%%%%%%%%%%%%%%%%%%%%%%%%%%%%%%%%%%%%%%%%%%%%%%%%%%%%%%%%%%%%%%%%%
%%%%%%%%%%%%%%%%%%%%%%%%%%%%%%%%%%%%%%%%%%%%%%%%%%%%%%%%%%%%%%%%%%%%%%%%%%%%%%%
% APPENDIX
%%%%%%%%%%%%%%%%%%%%%%%%%%%%%%%%%%%%%%%%%%%%%%%%%%%%%%%%%%%%%%%%%%%%%%%%%%%%%%%
%%%%%%%%%%%%%%%%%%%%%%%%%%%%%%%%%%%%%%%%%%%%%%%%%%%%%%%%%%%%%%%%%%%%%%%%%%%%%%%
\newpage
\appendix
\onecolumn


% You can have as much text here as you want. The main body must be at most $8$ pages long.
% For the final version, one more page can be added.
% If you want, you can use an appendix like this one, even using the one-column format.
% %%%%%%%%%%%%%%%%%%%%%%%%%%%%%%%%%%%%%%%%%%%%%%%%%%%%%%%%%%%%%%%%%%%%%%%%%%%%%%%
% %%%%%%%%%%%%%%%%%%%%%%%%%%%%%%%%%%%%%%%%%%%%%%%%%%%%%%%%%%%%%%%%%%%%%%%%%%%%%%%

\section{Definitions and Proofs}
  
\subsection{Partial exchangeability}\label{subsec:app-partial-exchange}
Here, we state a more formal version of Assumption~\ref{assumption:fl-exchangeable}.

\paragraph{Assumption~\ref{assumption:fl-exchangeable}} Define $S_i^k$ to be the $i$-th score on client $k$ i.e., $S_{1}^k := S(X_{1}^{k}, Y_{1}^{k})$, and $\sigma(\cdot)$ to mean the joint probability density function of random variables. Then, we assume there exist a probability vector $\lambda \in \Delta^K$, and random variables $\{S_{n_k + 1}^k\}_{k=1, \dots, K}$ such that we can decompose the probability density as
\[ \sigma(S(X_{\text{test}}, Y_{\text{test}})) = \sum_k \lambda_k \cdot \sigma(S_{n_k + 1}^k) \,,\]
and further, the joint probability is group-wise permutation invariant i.e., for any set of permutations $\{\pi_1, \dots, \pi_K\}$,
\begin{equation*}
\sigma\left(\begin{bmatrix}
    S_1^1 & \ldots & S_{n_1}^1 & S_{n_1 + 1}^1 \\
    \vdots & \ddots & \vdots & \vdots \\
    S_1^K & \ldots & S_{n_K}^K & S_{n_K + 1}^K \\
  \end{bmatrix}\right) = \;
\sigma\left(\begin{bmatrix}
    S_{\pi_1(1)}^1 & \ldots & S_{\pi_1(n_1)}^1 & S_{\pi_1(n_1 + 1)}^1 \\
    \vdots & \ddots & \vdots & \vdots \\
    S_{\pi_K(1)}^K & \ldots & S_{\pi_K(n_K)}^K & S_{\pi_K(n_K + 1)}^K \\
  \end{bmatrix}\right)\,.
\end{equation*}

\begin{remark}
    Consider the example where the client data is sampled independently from $(X_i^k, Y_i^k) \sim \mathbb{P_k}$ for $i=1, \dots, n_k$. Further suppose that the test point is independently drawn from $(X_{\text{test}},Y_{\text{test}}) \sim \sum_k \lambda_k \mathbb{P}_k$. We will show that this satisfies Assumption~\ref{assumption:fl-exchangeable}. Note that the scores on client $k$, $\,\{S_1^k, \dots, S_{n_k}^k\}$ are all IID and exchangeable with each other within client $k$. Now, denote $S^k_{n_k + 1} := S(X_{\text{test}},Y_{\text{test}}) | (X_{\text{test}},Y_{\text{test}}) \sim \mathbb{P}_k$. Then, clearly we have $\{S^k_1, \dots, S^k_{n_k + 1}\}$ are IID and hence exchangeable. Finally, note that $S(X_{\text{test}},Y_{\text{test}}) = S_{n_k + 1}^k$ with probability $\lambda_k$.
\end{remark}

\subsection{Proof of Theorem~\ref{theorem:fl-conformal}}\label{subsec:proof-of-main-theorem}
\begin{proof}
We denote the total number of samples for conformal calibration as $N=\sum_{k=1}^{K} n_k$. Given $\lambda_k \propto (n_k + 1)$ and $\sum_{k=1}^{K}\lambda_k = 1$, we have $\sum_{k=1}^{K}(n_k+1) = N+K$. Therefore,  
\begin{equation}\label{eq:thm-eq1}
\frac{\lambda_k}{n_k + 1} = \frac{1}{N+K}.
\end{equation}

Meanwhile, for each client $k$, we define $$m_k(q) :=|\{S(X_{i}^{k}, Y_{i}^{k})\leq q\}|\,.$$
Recall that we pick $\hat q_\alpha$ as the $\lceil(1 - \alpha)(N+K)\rceil$-th largest score i.e. it satisfies
\begin{equation}\label{eq:thm-eq2}
\sum_{k\in[K]} m_k(\hat q_\alpha) = \lceil(1 - \alpha)(N+K)\rceil\,. 
\end{equation}

Next, we define the event $\mathcal{E}$ as
\begin{equation}\label{eq:thm-event-def}
    \mathcal{E} = \left\{\forall k \in [K], \exists\,  \pi_k, (S^{k}_{\pi_k(1)}, \cdots, S^{k}_{\pi_k(n_k)}, S^{k}_{\pi_k(n_{k}+1)}) = (s^{k}_1, \cdots, s^{k}_{n_k}, s^{k}_{n_{k}+1}) \right\},
\end{equation}
where $\{s^{k}_i\}_{i \in [n_k + 1], k\in[K]}$ are the order statistics of the scores i.e., they represent the sorted numerical values of the score values. Note that the index assignment of these values to a particular score is still random and is unconditioned. Upon conditioning of the order statistics, the quantity $m_k(\hat q_\alpha)$ is a deterministic quantity.

Then we have 
\begin{equation}\label{eq:thm-eq3}
\begin{aligned}
&\mathbf{P}\left(S(X_{\text{test}}, Y_{\text{test}}) \leq \hat{q}_{\alpha} \,|\, \mathcal{E}\right) \\
= &\sum_{k=1}^{K} \lambda_k \cdot \mathbf{P}\left(S(X_{\text{test}}, Y_{\text{test}}) \leq \hat{q}_{\alpha} \,\big|\, \left\{S(X_{1}^{k}, Y_{1}^{k}), \, \dots, \, S(X_{n_k}^{k}, Y_{n_k}^{k}), \, S(X_{\text{test}}, Y_{\text{test}})\right\} \text{ are exchangeable}, \mathcal{E} \right)\\
\geq\, &\sum_{k=1}^{K}\lambda_k \cdot \frac{m_k(\hat q_\alpha)}{n_k+1}\\
=\, &\frac{\sum_{k=1}^{K} m_k(\hat q_\alpha)}{N+K}\\
=\, &\frac{\lceil (1-\alpha) (N+K) \rceil}{N+K}\\
% \geq\, &\frac{(1-\alpha)(nK+1)}
% {K(n+1)}\\
\geq\, &(1-\alpha),
\end{aligned}
\end{equation}
where we apply the partial exchangeable assumption (Assumption~\ref{assumption:fl-exchangeable}) for the first equality, and the first inequality is by exchangeability given $S(X_{1}^{k}, Y_{1}^{k}), \, \dots, \, S(X_{n}^{k}, Y_{n}^{k}), \, S(X_{\text{test}}, Y_{\text{test}})$ are exchangeable random variables.  The second equality is because of Eq.~\eqref{eq:thm-eq1}. 

Since Eq.~\eqref{eq:thm-eq3} holds for any $(s^{k}_1, \cdots, s^{k}_{n_k}, s^{k}_{n_{k}+1})$, $k\in[K]$, we can take expectation on both sides w.r.t. the order statistics and using the towering property of expectations we have 
\begin{equation}\label{eq:thm-eq3-uncondition}
\begin{aligned}
\mathbf{P}\left(S(X_{\text{test}}, Y_{\text{test}}) \leq \hat{q}_{\alpha}\right) 
\geq\, (1-\alpha).
\end{aligned}
\end{equation}

Next, we prove the upper bound of $\mathbf{P}\left(S(X_{\text{test}}, Y_{\text{test}}) \leq \hat{q}_{\alpha}\right)$. Suppose the ${\lceil (1-\alpha) (N+K) \rceil}$ largest value in $\left\{S\left(X_{i}^{k}, Y_{i}^{k}\right)\right\}_{i\in [n_k],\, k\in[K]}$ is from the $\hat{k}$-th client,  then we have
\begin{equation}\label{eq:thm-eq4}
\begin{aligned}
&\mathbf{P}\left(S(X_{\text{test}}, Y_{\text{test}}) \leq \hat{q}_{\alpha} \,|\, \mathcal{E}\right) \\
\leq\, &\sum_{k=1}^{K}\frac{\mathbb{1}\{\hat{k}\neq k\}\cdot\lambda_k\cdot (m_k(\hat q_\alpha)+1) + \mathbb{1}\{\hat{k}=k\}\cdot\lambda_k \cdot m_k(\hat q_\alpha)}{n_k+1}\\
=\, &\frac{(K-1) + \sum_{k=1}^{K} m_k(\hat q_\alpha)}{N+K}\\
=\, &\frac{(K-1)+\lceil (1-\alpha) (N+K) \rceil}{N+K}\\
\leq \, &(1-\alpha) + \frac{K}{N+K},
\end{aligned}
\end{equation} \vspace{-1em}
where the second equality is due to Eq~\eqref{eq:thm-eq2}, and this concludes our proof.
\end{proof}

\begin{figure*}[t!]
\centering
\begin{subfigure}{.6\textwidth}
  \includegraphics[width=\linewidth]{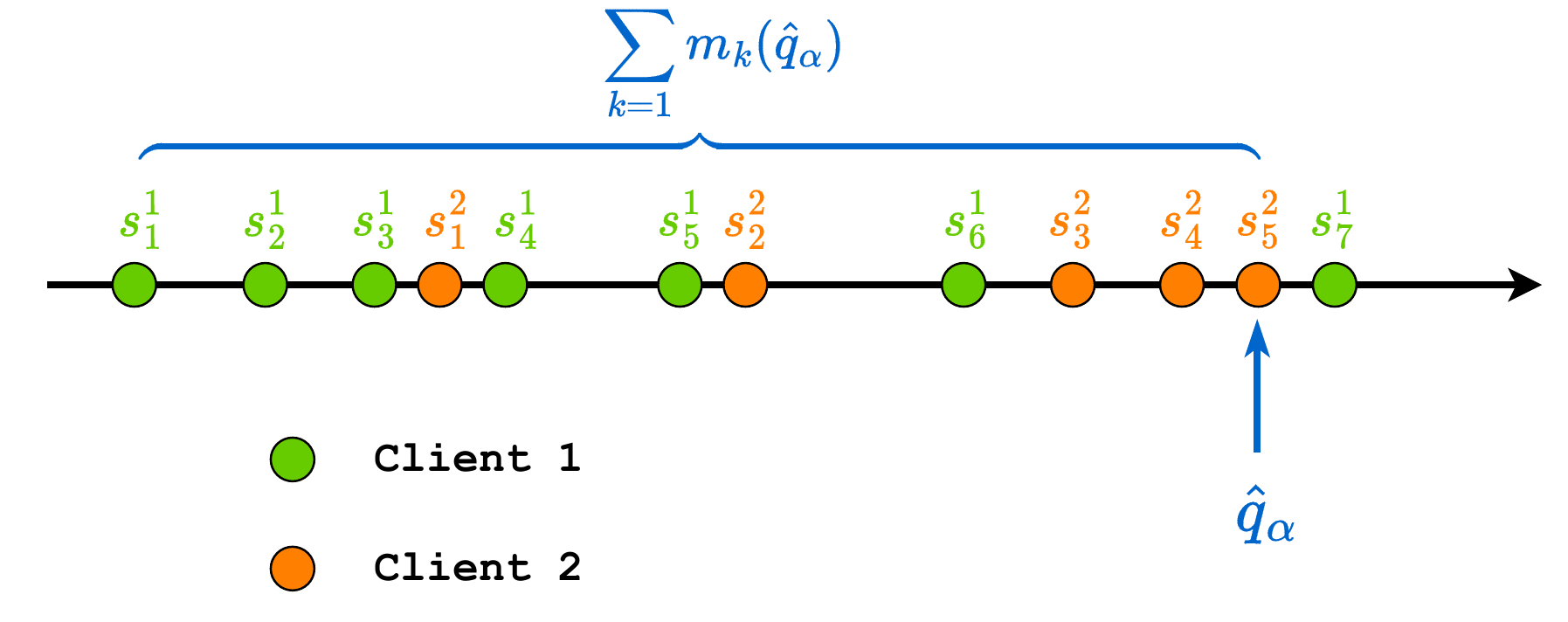}
\end{subfigure}
\vspace{-0.1in}
    \caption{Visualization of the $\hat{q}_{\alpha}$ of Algorithm~\ref{alg:fcps} with two clients and $\alpha=0.3$, where the first client has seven calibration points and the second client has five calibration points.}
    \label{fig:appendix-compute-q-aht}
    \vskip -0.1in
\end{figure*}

\subsection{Proof of Corollary~\ref{corollary:fl-conformal-approx}}
\begin{proof}
To begin with, we first use $\tilde{q}_{\alpha}$ to denote $\varepsilon$-approximate $(1-\alpha)$-quantile. 
Then similar to  Eq.~\eqref{eq:thm-eq3} in Theorem~\ref{theorem:fl-conformal}, the lower bound of $\mathbf{P}\left(S(X_{\text{test}}, Y_{\text{test}}) \leq \tilde{q}_{\alpha}\right)$ is
\begin{equation}
   \mathbf{P}\left(S(X_{\text{test}}, Y_{\text{test}}) \leq \tilde{q}_{\alpha}|\,\mathcal{E}\,\right) 
   \geq\, \frac{\sum_{k=1}^{K} m_k(\tilde q_\alpha)}{N+K}
   % \geq\, \frac{\lfloor (1-\alpha) (N+K) \rfloor}{N+K}  
   \geq\, \frac{ (1-\alpha -\varepsilon) (N+K) -1 }{N+K},
\end{equation}
where the second inequality is by the Definition~\ref{def:approx-quantile} and $\mathcal{E}$ is defined in Eq.~\eqref{eq:thm-event-def}. Similarly, based on Eq.~\eqref{eq:thm-eq4}, we can prove the upper bound of $\mathbf{P}\left(S(X_{\text{test}}, Y_{\text{test}}) \leq \tilde{q}_{\alpha}\right)$.
\end{proof}

\section{Federated Conformal Prediction with Unknown Weights}
\subsection{Extension to Unknown Weights}\label{subsec:appendix-extension-to-unkown-lambda}
Suppose that Assumption~\ref{assumption:fl-exchangeable} holds for some arbitrary an \emph{unknown} probability vector $\lambda \in \Lambda \subseteq \Delta^K$. While we may not know the exact weight $\lambda$, we know the set $\Lambda$ it could belong to. Thus, $\Lambda$ measures our uncertainty about the true weight $\lambda$.

Weighted quantile: Pick $\hat q_\alpha$ such that
\begin{equation}\label{eq:unknown_weighted_qunatile}
\hat q_\alpha = \min \left\{q \,\Big|\, \min_{\hat\lambda \in \Lambda}\sum_{k=1}^{K}\frac{\hat\lambda_k m_k(q)}{n_k+1} \geq 1-\alpha \right\}\,.    
\end{equation}

 From Eq.~\eqref{eq:thm-eq3} in the proof of Theorem~\ref{theorem:fl-conformal}, we have almost surely that
\begin{align*}
\mathbf{P}\left(S(X_{\text{test}}, Y_{\text{test}}) \leq \hat q_\alpha \right) &\geq\, \sum_{k=1}^{K}\frac{\lambda_k m_k(\hat q_\alpha)}{n_k+1}    \\
&\geq\, \min_{\hat\lambda \in \Lambda}\sum_{k=1}^{K}\frac{\hat\lambda_k m_k(\hat q_\alpha)}{n_k+1}\\
&\geq (1-\alpha)\,.
\end{align*}
The second inequality above followed from the assumption that the true $\lambda \in \Lambda$, while the final inequality used the definition of $\hat q_\alpha$. We will next consider some important special cases.

\paragraph{Example 1.} Consider the case where we have $\lambda_k \approx \frac{n_k + 1}{N + K}$. More specifically, suppose there exist $\delta \in [0, 1)$ such that,
\begin{equation}\label{eq:appendix-approx-lambda}
    \lambda_k \geq (1 - \delta)\frac{n_k + 1}{N+K} \quad \forall k \in [K]\,.
\end{equation}

This is an approximate version of the setting in Theorem~\ref{theorem:fl-conformal} where $\delta$ controls the approximation factor. Let us construct the set 
\[ \Lambda = \left\{\hat \lambda \,\big|\, \hat\lambda_k \geq (1 - \delta)\tfrac{n_k + 1}{N+K}\right\}\,.
\]
Since $\lambda \in \Lambda$, we have
\begin{align*}
\mathbf{P}\left(S(X_{\text{test}}, Y_{\text{test}}) \leq q \right) &\geq\, \min_{\hat\lambda \in \Lambda}\sum_{k=1}^{K}\frac{\hat\lambda_k m_k(q)}{n_k+1}\\
&= \min_{\hat\lambda \in \Lambda}\sum_{k=1}^{K}\frac{(\hat\lambda_k - \tfrac{n_k+1}{N +K}) m_k(q)}{n_k+1} +  \frac{( \tfrac{n_k+1}{N +K}) m_k(q)}{n_k+1}\\
&= \min_{\hat\lambda \in \Lambda}\sum_{k=1}^{K}\left( \tfrac{\hat\lambda_k (N+K)}{n_k + 1} - 1 \right) \frac{m_k(q)}{N + K} + \frac{\sum_k m_k(q)}{N+ K}.
\end{align*}
Now, by the construction of $\Lambda$, we can lower bound the right-hand side above, proving
\begin{equation}\label{eq:appendix-approx-lambda-2}
\begin{aligned}
\min_{\hat\lambda \in \Lambda}\sum_{k=1}^{K}\frac{\hat\lambda_k m_k(q)}{n_k+1} &\geq\,   \sum_{k=1}^{K}\left( 1 - \delta - 1 \right) \frac{m_k(q)}{N + K} + \frac{\sum_k m_k(q)}{N+ K} \\
&= (1 - \delta) \frac{\sum_k m_k(q)}{N+ K} \,.
\end{aligned}
\end{equation}
This shows that using the $\lceil (N + K)(1 - \alpha)/(1-\delta) \rceil$-th largest score as $\hat q_\alpha$ suffices to give provable $(1-\alpha)$ coverage. This neatly recovers the algorithm as well as guarantee when $\lambda_k = \tfrac{n_k + 1}{N+K}$ in Theorem~\ref{theorem:fl-conformal} by setting $\varepsilon = 0$. Note that using Eq.~\eqref{eq:unknown_weighted_qunatile} directly would also yield the required $(1-\alpha)$ coverage with a smaller value of $\hat q_\alpha$ (and hence smaller set sizes). However, this would be less interpretable.

\paragraph{Example 2.} Next, consider the special case where we know that $\lambda_k = \frac{1}{K}$ i.e. we want to weight every client equally. In this case, we have
\[
\frac{1}{K}\sum_{k=1}^{K}\frac{m_k(\hat q_\alpha)}{n_k+1}  \leq \mathbf{P}\left(S(X_{\text{test}}, Y_{\text{test}}) \leq \hat q_\alpha \right) \leq\, \frac{1}{K}\sum_{k=1}^{K}\frac{m_k(\hat q_\alpha) + 1}{n_k + 1}\,.
\]
The gap between the upper and lower bounds, in this case, is $\frac{1}{K}\sum_{k=1}^{K}\frac{1}{n_k + 1}$. This is the inverse of the \emph{harmonic mean} of $\{(n_1 +1), \dots, (n_K +1)\}$. Thus,
\[
\max_{k} \frac{1}{n_k+1} \geq \frac{1}{K}\sum_{k=1}^{K}\frac{1}{n_k + 1} \geq  \frac{K}{\sum_k (n_k +1)} = \frac{1}{N/K + 1}\,,
\]
with equalities holding only if all clients have equal data with $n_k = \tfrac{K}{N}$. Thus, the gap when $\lambda_k = \tfrac{1}{K}$ is larger than when we have $\lambda_k \propto (n_k+1)$ and is more sensitive to clients with little data. However, it is still better than using a client's data on its own, which would have a gap of $\max_{k} \frac{1}{n_k+1}$.

\subsection{Unified Analysis of Federated Conformal Method}
In this subsection, we present a unified analysis of our proposed conformal prediction by considering the following three factors: (a) $K$ number of clients; (b) $\delta$ error parameter of $\lambda$; (c) $\varepsilon$ error parameter of distributed quantile estimation.

\begin{theorem}\label{theorem:appendix-fl-conformal-unify}
Suppose there are $n_k$ samples from the $k$-th client and $\lambda_k \propto (n_k + 1)$, if we define $\hat{q}_{\alpha}$ as the ${\lceil (1-\alpha) (N+1) \rceil}$ largest value in $\{(S(X_{i}^{k}, Y_{i}^{k}))\}_{i\in [n_k],\, k\in[K]}$. Then for $(X_{\text{test}}, Y_{\text{test}}) \sim \mathbb{Q}_{\lambda}$, we have
\begin{equation}\label{eq:theorem-results-appendix}
    \mathbb{E}\left[\mathbb{1}\{Y_{\text{test}} \in \, \mathcal{C}_\alpha(X_{\text{test}}) \}\right] \geq (1-\delta)\frac{(1 - \alpha - \varepsilon)(N+1) - \mathbb{1}\{\varepsilon > 0\}}{N+K},
\end{equation}
where $\delta \in [0, 1)$ is the error parameter of $\lambda$ defined in Eq.~\eqref{eq:appendix-approx-lambda}, $\varepsilon \in [0, 1)$ is the error parameter of distributed quantile estimation defined in Definition~\ref{def:approx-quantile}.
\end{theorem}

\begin{proof}
To start with, by Eq.~\eqref{eq:appendix-approx-lambda-2}, we have 
\begin{equation}
\min_{\hat\lambda \in \Lambda}\sum_{k=1}^{K}\frac{\hat\lambda_k m_k(\hat{q}_{\alpha})}{n_k+1} \geq (1 - \delta) \frac{\sum_k m_k(\hat{q}_{\alpha})}{N+ K} \,,
\end{equation}
then by applying the Definition~\ref{def:approx-quantile}, we have 
\begin{equation}
    \sum_k m_k(\hat{q}_{\alpha}) \geq (1-\alpha - \varepsilon)(N+1) - \mathbb{1}\{\varepsilon>0\},
\end{equation}
therefore, we have 
\begin{equation}
    \mathbbm{P}\left(S(X_{\text{test}}, Y_{\text{test}}) \leq  \hat{q}_{\alpha} \right) \geq  \min_{\hat\lambda \in \Lambda}\sum_{k=1}^{K}\frac{\hat\lambda_k m_k(\hat{q}_{\alpha})}{n_k+1} \geq (1-\delta)\frac{(1 - \alpha - \varepsilon)(N+1) - \mathbb{1}\{\varepsilon > 0\}}{N+K},
\end{equation}
which concludes our proof.
\end{proof}
\begin{remark}
When $\{(S(X_{i}^{k}, Y_{i}^{k}))\}_{i\in [n_k],\, k\in[K]}$ and $S(X_{\text{test}}, Y_{\text{test}})$ are exchangeable, then it is equivalent to the setting where $K=1$. In this scenario, when $\delta = \varepsilon = 0$, Eq.~\eqref{eq:theorem-results-appendix} is the same as the standard coverage guarantee of non-FL conformal prediction methods.
\end{remark}

\section{Additional Results of Distributed Quantile Estimate}\label{sec:appendix-dist-quantile}

\subsection{Communication required}
In this section, we take the algorithm from \citet{huang2011sampling} as an example and provide a detailed theoretical statement below.

As shown in Theorem 3 of \citet{huang2011sampling}, computing an $\varepsilon$-approximate  $(1-\alpha)$-quantile requires at most $O(K/\varepsilon)$ bits to be communicated to the server. Therefore, incorporating this guarantee, we can restate Corollary~\ref{corollary:fl-conformal-approx}. 

\paragraph{Example 1.} Suppose we have $K$ clients, and there are $n_k$ samples from the $k$-th client and $\lambda_k = (n_k + 1) / (N+K)$, where $N=\sum_{k=1}^{K}n_k$ is the total number of calibration samples. 
Suppose $N\geq K$ and $\alpha < 1/(N+K)$. Then, let us set $\varepsilon=((N+K)\alpha - 1)/N$. For this value of $\varepsilon$, the distributed quantile estimation algorithm takes $O(K/\varepsilon)$ bits of communication and guarantees that the output $C_{\alpha}$ of Algorithm~\ref{alg:fcps} satisfies $\mathbf{P}\left(Y \in \, \mathcal{C}_\alpha(X)\right) \geq 1 - 2\alpha$.

\subsection{Differential privacy}
While the default Federated Learning (FL) setup does not provide formal privacy guarantees, we can extend our FCP framework to incorporate differential privacy (DP). In particular, we sketch how to adapt our existing inexact-quantile guarantees, specifically Corollary~\ref{corollary:fl-conformal-approx}, to infer coverage under DP.

We leverage the exponential mechanism for selecting a private quantile~\cite{mcsherry2007mechanism}. Assume we are given a set of scores $(s_1, \dots, s_N)$ and a privacy parameter $\delta$ is chosen (while typically $\varepsilon$ is used to indicate the privacy parameter, here we use $\delta$ to avoid notation clash with $\epsilon$ from Corollary~\ref{corollary:fl-conformal-approx}). Define the following utility function:
\[
    u(x) = -\left|{| \{i: \text{ s.t. } s_i \leq x\}|} - N(1-\alpha)\right|\,, \quad \text{ defined for } x \in (s_1, \dots, s_N)\,.
\]
Its sensitivity can clearly be seen to be $1$. Thus, to achieve $\delta$-DP, we need to output a quantile $x$ such that 
\[
    \Pr(x = s_i) \propto \exp(\delta u(s_i) / 2)\,.
\]
This is equivalent to the following procedure:
\begin{enumerate}
    \item Pick $r \in [N]$ such that $\Pr(r = i) \propto \exp(- {\delta}{|i - (1-\alpha)N|} / 2)$ 
    \item Return the $r$-th largest number in $\{s_1, \dots, s_N\}$ using an exact distributed quantile estimator.
\end{enumerate}
Thus, by the exponential mechanism~\citep{mcsherry2007mechanism}, the output of the above mechanism satisfies $\delta$-DP. Further, by examining the tail of the exponential distribution, we have that $\beta \in [1-\alpha \pm O(\tfrac{-\log(\delta\alpha)}{\delta N})]$ with probability at least $1 - \alpha$. This satisfies the $\epsilon$ approximate quantile estimator for $\epsilon=\tfrac{1}{\delta N}$ as in Definition~\ref{def:approx-quantile}. Combining these results with Corollary~\ref{corollary:fl-conformal-approx}, we get the following result.

\begin{corollary}
The exponential mechanism returns a quantile-estimator that satisfies $\delta$-DP and has a coverage guarantee of
    \begin{align*}
    \mathbf{P}\left(Y \in \, \mathcal{C}_\alpha(X)\right) \geq 
    1 - 2\alpha - O\left(\frac{\log(\nicefrac{1}{\delta \alpha})}{\delta N}\right) - \frac{\mathbb{1}\{\varepsilon > 0\}}{N+K}\,.
    \end{align*}
\end{corollary}

A similar coverage guarantee can be obtained with stronger \emph{local} DP guarantees by relying on the privately distributed quantile estimator such as those in \citet{pillutla2022differentially}.

\section{Conformity Score Functions}
\label{app:conformal-score-functions}
    Given a trained classifier $f$, a score function $S$, and an exchangeable calibration set of features $X \in \mathbb{R}^d$ with labels $Y \in  \mathcal{Y} = \{1, 2, \ldots, J\}$, a prediction set that outputs a subset of classes $2^\mathcal{Y}$ can be formed by 
    \begin{equation}
        % \small
        \mathcal{C}(X) = \left\{y \in \mathcal{Y}: S(X, y) \leq \hat{q}_\alpha \right\},
    \end{equation}
    where $\hat{q}_\alpha$ is the $(1-\alpha)$ quantile of the calibration scores ${ \hat{q}_\alpha = \text{Quantile}\left(\{s_1, s_2, \ldots, s_n\}, \frac{\lceil{(n+1)(1-\alpha)\rceil}}{n}\right) }$.
    We consider three commonly used score functions for conformal prediction classification tasks in our experiments: 
    \begin{enumerate}
        \item \textit{least ambiguous set-valued classifiers} (LAC)~\cite{doi:10.1080/01621459.2017.1395341}
        \begin{equation}
            % \small
            S_\text{LAC}(x, y) = 1 - \left[f(x)\right]_{y}, 
        \end{equation}
    where $\left[f(x)\right]_y$ indexes the score of the true label,
        \item \textit{adaptive prediction sets} (APS)~\cite{NEURIPS2020_244edd7e}
    \begin{equation}
        % \small
        S_\text{APS}(x, y) = \sum_{j=1}^{J^\prime} \left[\pi\left(f(x)\right)\right]_{j},
    \end{equation}
    where $ { J^\prime = \sup\left\{j \in \mathcal{Y}: \sum_{i=1}^j \left[\pi(f(x))\right]_{i} \leq 1 - \alpha \right\} }$ and $\pi\left(f(x)\right)$ is the permutation that sorts the scores in descending order, 
    \item and \textit{regularized adaptive prediction sets} (RAPS)~\cite{angelopoulos2020sets}
    \begin{equation}
        % \small
         S_\text{RAPS}(x, y) = \sum_{j=1}^{J^\prime} \left[\pi\left(f(x)\right) + a \cdot \mathbb{1}\left[\left\{j > b\right\}\right] \right]_{j},
    \end{equation}
    where $\mathbb{1}\{\cdot\}$ is the indicator function, $J^\prime$ is defined same as above, and $(a, b)$ are regularization parameters.
    \end{enumerate}

\newpage
\section{Dataset Information}
\label{app:data}

\begin{figure}[ht]
    \vskip 0.2in
    \begin{center}
    \centerline{\includegraphics[width=0.8\columnwidth]{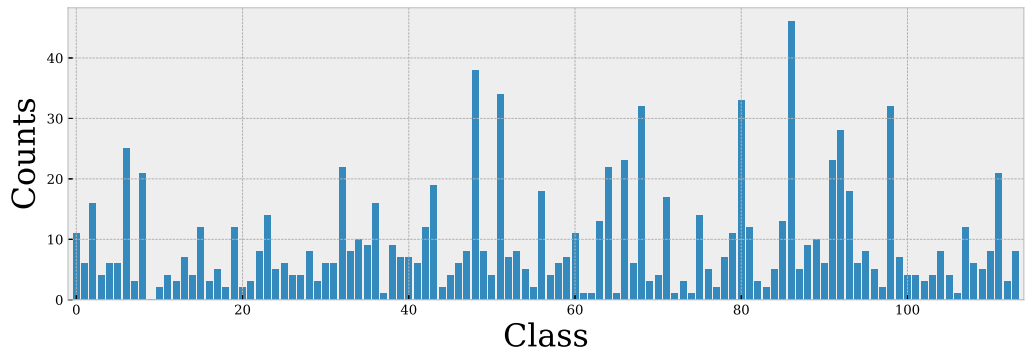}}
    \caption{\textbf{Class distribution of Fitzpatrick17k skin lesion dataset.} The Fitzpatrick17K dataset contains 16,577 photography images collected from two dermatology atlases with labels for 114 different skin conditions.}
    \label{fig:skin-class-dist}
    \end{center}
    \vskip -0.25in
\end{figure}

    \begin{figure}[ht]
    \vskip 0.2in
    \begin{center}
    \centerline{\includegraphics[width=0.7\columnwidth]{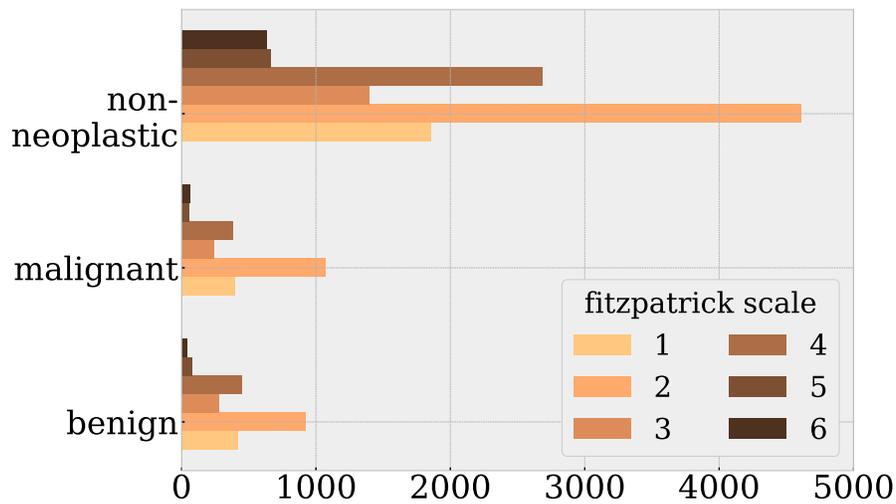}}
    \caption{\textbf{Distribution of skin types of Fitzpatrick17k skin lesion dataset}. Most images in the Fitzpatrick17k are also labeled with their Fitzpatrick skin type, which measures the amount of melanin pigment in the skin. 
}
    \label{fig:skin-rate}
    \end{center}
    \vskip -0.3in
\end{figure}

\begin{figure}[ht]
    % \vskip 0.2in
    \begin{center}
    \centerline{\includegraphics[width=0.6\columnwidth]{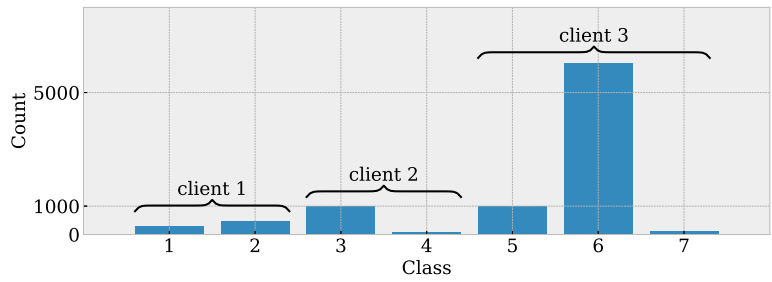}}
    \caption{\textbf{Class distribution and client partition on DermaMNIST dataset.} DermaMNIST is a dermatoscopy dataset with 10,015 images labeled with one of 7 conditions.}
    \label{fig:derma}
    \end{center}
    \vskip -0.2in
\end{figure}

\begin{figure}[ht]
    % \vskip 0.2in
    \begin{center}
    \centerline{\includegraphics[width=0.6\columnwidth]{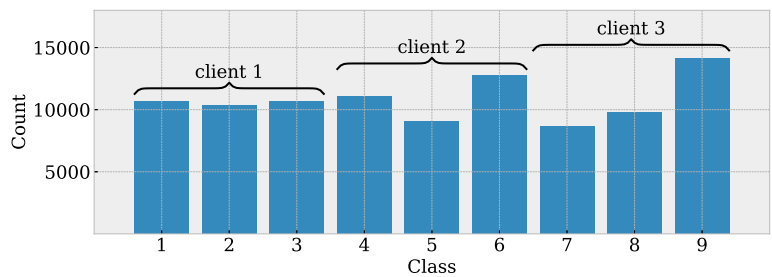}}
    \caption{\textbf{Class distribution and client partition on PathMNIST dataset.} PathMNIST is a colon pathology dataset with 107,180 images labeled with one of 9 conditions.}
    \label{fig:path}
    \end{center}
    \vskip -0.2in
\end{figure}

\begin{figure}[ht]
    \vskip 0.2in
    \begin{center}
    \centerline{\includegraphics[width=0.6\columnwidth]{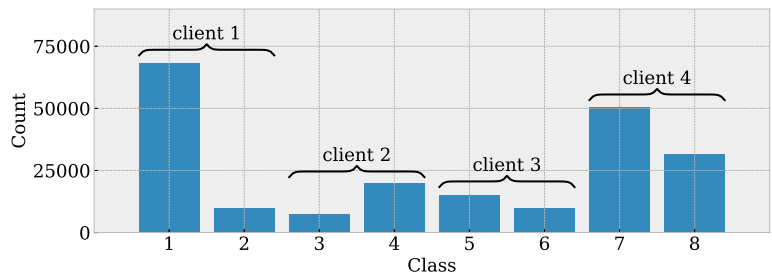}}
    \caption{\textbf{Class distribution and client partition on TissueMNIST dataset.} TissueMNIST is a kidney cortex microscope dataset of 236,386 images labeled with one of 8 conditions.}
    \label{fig:tissue}
    \end{center}
    \vskip -0.2in
\end{figure}

\clearpage
\section{Implementation Details}
\label{app:details}
For model architectures, we used LeNet~\cite{lecun1998gradient} for FashionMNIST, Efficient-Net-B1~\cite{tan2019efficientnet} pretrained on ImageNet~\cite{deng2009imagenet} for Fitzpatrick17k, and ResNet-14 (based off of \citet{Idelbayev18a} implementation) for the rest of the datasets.

We use the TCT approach proposed by \citet{yu2022tct} to train a distributed model suited for data heterogeneity. 
TCT has two optimization stages: a pretraining stage that trains a FedAvg model and a second stage that solves a convex approximation of the trained model.
We found the squared loss used in the paper resulted in poor calibration and instead used a cross-entropy loss in the second stage.
For FedAvg, 200 communication rounds with five local epochs.
For stage 2 of TCT, we take the FedAvg model trained with 100 epochs (instead of 200) and additionally train 100 communication rounds with a learning rate of 0.0001 and 500 local steps.

For the Fitzpatrick17k dataset, we use the Torchvision implementation of the EfficientNet architecture pre-trained on Imagenet-V1.
For all other datasets, we use ResNet-14.
We train all models with SGD with 0.9 momentum, a learning rate of 0.01 (0.001 for Fitzpatrick17k), and a minibatch size of 64.
Data augmentation, such as random flipping and cropping, was applied to all datasets during training; for the Fitzpatrick17k dataset, random color jittering and rotations were also applied.

For RAPS, we set the regularization parameters $a=1$ and $b=0.001$ ($b=0.00001$ for the Fitzpatrick17k dataset).

\section{Additional Experimental Results}
\label{app:experiments}

\begin{table*}[t]
    \caption{\textbf{Results of conformal prediction with different models.} We report coverage and size of prediction sets calibrated centrally and with our decentralized framework (FedAvg, FedProx, and TCT). The mean over ten random trials is reported with standard deviations in the range of $\pm 0.01$ for coverage results and $\pm 0.1$ for size. Smaller set sizes indicate more efficient prediction sets.}
    \label{tab:full-results}
    \vskip 0.15in
    \begin{center}
    \begin{small}
    \begin{sc}\resizebox{0.98\textwidth}{!}{%
    \begin{tabular}{lcc|cc|cc|cc|cc}
    \toprule
    & & & \multicolumn{2}{c|}{FedAvg} & \multicolumn{2}{c|}{FedProx} & \multicolumn{2}{c|}{TCT} & \multicolumn{2}{c}{Centralized} \\
    Dataset & classes & $1 - \alpha$ & coverage & size & coverage & size & coverage & size & coverage & size\\
    \midrule
    \multirow{2}{2.9cm}{FashionMNIST} & \multirow{2}{1cm}{10} 
        & $0.90$ & $0.90$ & $3.2 $ & $0.90 $ & $2.2$ & $0.91 $ & $1.2 $ & $0.90$ & $1.1$ \\
        & & $0.80$ & $0.80$ & $2.2 $ & $0.80 $ & $1.5$ & $0.89 $ & $1.1 $ & $0.88$ & $1.0$ \\
        & & $0.70$ & $0.70$ & $1.3 $ & $0.70 $ & $1.2$ & $0.89 $ & $1.0 $ & $0.88$ & $1.0$ \\
    \midrule
    \multirow{2}{2.9cm}{CIFAR-10} & \multirow{2}{1cm}{10} 
        & $0.90$ & $0.90 $ & $3.8 $ & $0.90 $ & $3.5 $ & $0.90 $ & $1.3 $ & $0.90 $ & $1.1 $\\
        & & $0.80$ & $0.80 $ & $2.5 $ & $0.80 $ & $2.3 $ & $0.82 $ & $1.0 $ & $0.88 $ & $1.0 $ \\
        & & $0.70$ & $0.70 $ & $1.8 $ & $0.70 $ & $1.6 $ & $0.82 $ & $1.0 $ & $0.88 $ & $1.0 $ \\
    \midrule
    \multirow{2}{2.9cm}{CIFAR-100} & \multirow{2}{1cm}{100} 
        & $0.90$ & $0.90$ & $17.3$ & $0.90$ & $17.9$ & $0.90$& $5.6$ & $0.90$ & $4.8$\\
        & & $0.80$ & $0.80$ & $9.2 $ & $0.80$ & $9.9$ & $0.80$& $2.8$ & $0.80$ & $2.4$\\
        & & $0.70$ & $0.70$ & $5.4 $ & $0.70$ & $6.1$ & $0.70$& $1.6$ & $0.70$ & $1.4$\\
    \midrule
    \multirow{2}{2.9cm}{DermaMNIST} & \multirow{2}{1cm}{7}
        & $0.90$ & $0.90 $ & $3.1 $ & $0.90 $ & $3.2 $ & $0.90 $& $2.4$ & $0.90$ & $1.7$\\
        & & $0.80$ & $0.81 $ & $2.1 $ & $0.80 $ & $2.2 $ & $0.81 $& $1.3 $ & $0.81$ & $1.2$\\
        & & $0.70$ & $0.73 $ & $1.5 $ & $0.71 $ & $1.5 $ & $0.74 $& $1.0 $ & $0.75$ & $1.0$\\
    \midrule
    \multirow{2}{2.9cm}{PathMNIST} & \multirow{2}{1cm}{9} 
        & $0.90$ & $0.90 $ & $3.2 $ & $0.90 $ & $2.8 $ & $0.90 $& $1.2$ & $0.90 $ & $1.0 $\\
        & & $0.80$ & $0.80 $ & $2.5 $ & $0.80 $ & $2.0 $ & $0.84 $& $1.0 $ & $0.89 $ & $1.0 $\\
        & & $0.70$ & $0.70 $ & $2.0 $ & $0.70 $ & $1.5 $ & $0.84 $& $1.0 $ & $0.89 $ & $1.0 $\\
    \midrule
    \multirow{2}{2.9cm}{TissueMNIST} & \multirow{2}{1cm}{8}
        & $0.90$ & $0.90 $ & $5.3 $ & $0.90 $ & $5.2 $ & $0.90 $& $2.7$ & $0.90 $ & $3.0 $\\
        & & $0.80$ & $0.80 $ & $4.2 $ & $0.80 $ & $4.1 $ & $0.80 $ & $1.8 $ & $0.80 $ & $2.0 $\\
        & & $0.70$ & $0.70 $ & $3.3 $ & $0.70 $ & $3.3 $ & $0.70 $& $1.3 $ & $0.70 $ & $1.5 $\\
    \midrule
    \multirow{2}{2.9cm}{\shortstack[l]{Fitzpatrick17K \\ (disease partition)}} & \multirow{2}{1cm}{114} 
        & $0.90$ & $0.91 $ & $24.9 $ & $0.91$ & $25.6$ & $0.91 $ & $20.0 $ & $0.90 $ & $16.1 $ \\
        & & $0.80$ & $0.79 $ & $8.6 $ & $0.80$ & $8.7$ & $0.80 $ & $6.0 $ & $0.81 $ & $6.1 $ \\
        & & $0.70$ & $0.69 $ & $3.8 $ & $0.69$ & $3.8$ & $0.70 $ & $2.8 $ & $0.71 $ & $2.7 $ \\
    \midrule
    \multirow{2}{2.9cm}{\shortstack[l]{Fitzpatrick17K\\(skin type partition)}} & \multirow{2}{1cm}{114}
        & $0.90$ & $0.91$ & $22.8$ & $0.91$ & $22.6$ & $0.91 $ & $21.3 $ & $0.90 $ & $18.2 $ \\
        & & $0.80$ & $0.81$ & $9.2$ & $0.81$ & $9.7$ & $0.80 $ & $7.4 $ & $0.80 $ & $6.3 $ \\
        & & $0.70$ & $0.69$ & $4.1$ & $0.72$ & $4.5$ & $0.70 $ & $3.2 $ & $0.70 $ & $2.6 $ \\
    \bottomrule
    \end{tabular}
    }
    \end{sc}
    \end{small}
    \end{center}
    \vskip -0.1in
\end{table*}

% \todo{check score function labels; move to appendix}
\begin{table}[t]
\small
\caption{\textbf{Efficiency gain of TCT is robust across the choice of the score function.} Comparing the mean size of prediction sets at $\alpha=0.1$ of CIFAR-100 test examples that were correctly predicted by top-1 across all three models (TCT, FedAvg, FedProx). All methods achieve perfect coverage on this subset.}
\label{tab:cifar-efficiency-score-compare}
\vskip 0.15in
\begin{center}
\begin{small}
\begin{sc}
\begin{tabular}{c|ccc}
\toprule
& LAC & APS & RAPS \\
method & size & size & size\\
\midrule
FedAvg  & $13.2$ & $13.3$ & $7.7$ \\
FedProx & $12.9$ & $14.2$ & $8.1$ \\
TCT     & $3.2$ & $7.6$  & $2.4$ \\
\bottomrule
\end{tabular}
\end{sc}
\end{small}
\end{center}
\vskip -0.1in
\end{table}

\begin{table*}[t]
\small
\caption{\textbf{RAPS is more adaptive than LAC and APS.} An adaptive conformal predictor outputs larger sets when the predictor is highly uncertain and smaller sets when the predictor is highly confident. We evaluated the size-conditional coverage and average set size of each quartile of set sizes and see that LAC has lower than $1-\alpha$ coverage on prediction with larger sizes. In comparison, RAPS has tighter $1-\alpha$ coverage across quartiles.}
\label{tab:size-conditional-coverage}
\vskip 0.15in
\begin{center}
\begin{small}
\begin{sc}
\begin{tabular}{cc|cc|cc|cc|cc}
\toprule
& size percentile & \multicolumn{2}{c|}{$0-25$} & \multicolumn{2}{c|}{$25-50$} & \multicolumn{2}{c|}{$50-75$} & \multicolumn{2}{c}{$75-100$}\\
\midrule
Dataset & \shortstack[c]{score \\ function} & coverage & size & coverage & size & coverage & size & coverage & size\\
\midrule
\multirow{3}{2cm}{\shortstack[l]{CIFAR-100}} 
&  LAC & $0.97$ & $2.0$ & $0.93$ & $4.5$ & $0.86$ & $6.9$ & $0.81$ & $10.1$\\
&  APS & $0.96$ & $2.1$ & $0.97$ & $8.4$ & $0.97$ & $19.3$ & $0.96$ & $37.6$\\
& RAPS & $0.94$ & $1.3$ & $0.92$ & $3.8$ & $0.89$ & $8.3$ & $0.88$ & $18.6$\\
\midrule
\multirow{3}{2cm}{\shortstack[l]{Fitzpatrick\\(Disease)}} 
&  LAC & $0.96$ & $2.5$ & $0.89$ & $9.7$ & $0.86$ & $20.7$ & $0.84$ & $37.8$\\
&  APS & $0.94$ & $7.0$ & $0.97$ & $45.6$ & $0.97$ & $74.5$ & $0.97$ & $89.5$\\
& RAPS & $0.92$ & $1.2$ & $0.87$ & $10.2$ & $0.90$ & $32.8$ & $0.92$ & $56.9$\\
% \midrule
% \multirow{3}{2cm}{\shortstack[l]{Fitzpatrick\\(Skin type)}} 
% & LAC & $0.94 $ & $0.82 $ & $0.76 $ \\
% & APS & $0.97 $ & $0.98 $ & $0.98 $ \\
% & RAPS & $0.92$ & $0.90 $ & $0.89 $ \\
\bottomrule
\end{tabular}
\end{sc}
\end{small}
\end{center}
\vskip -0.1in
\end{table*}       

\begin{figure}[ht]
 \begin{center}
  \centering
    \begin{subfigure}{.50\textwidth}
      \centering
      \includegraphics[width=\linewidth]{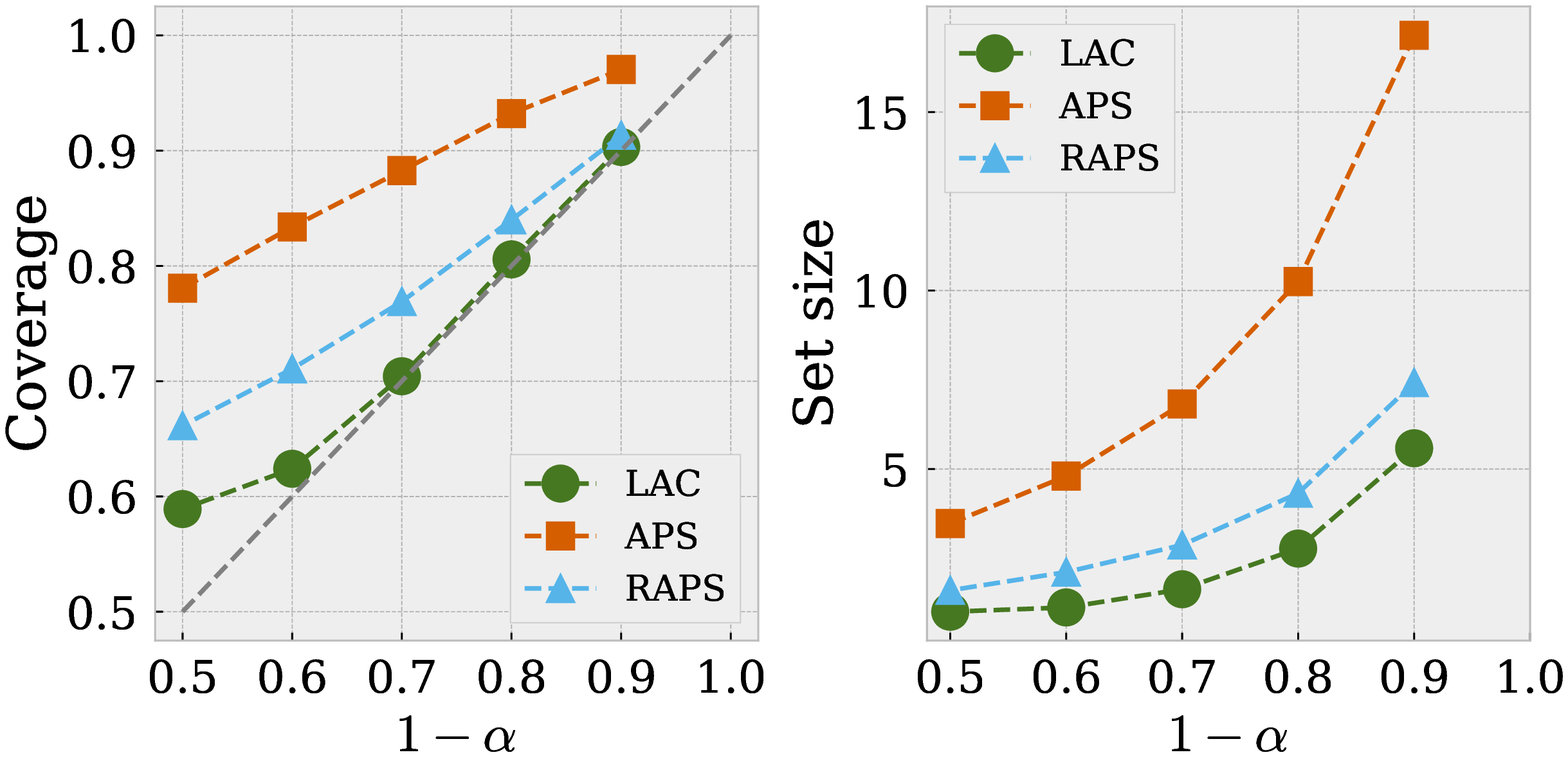}
      \caption{\textbf{CIFAR-100}}
      % \label{fig:challenge-exchangeability}
    \end{subfigure}\hfill
    \begin{subfigure}{.50\textwidth}
      \centering
      \includegraphics[width=\linewidth]{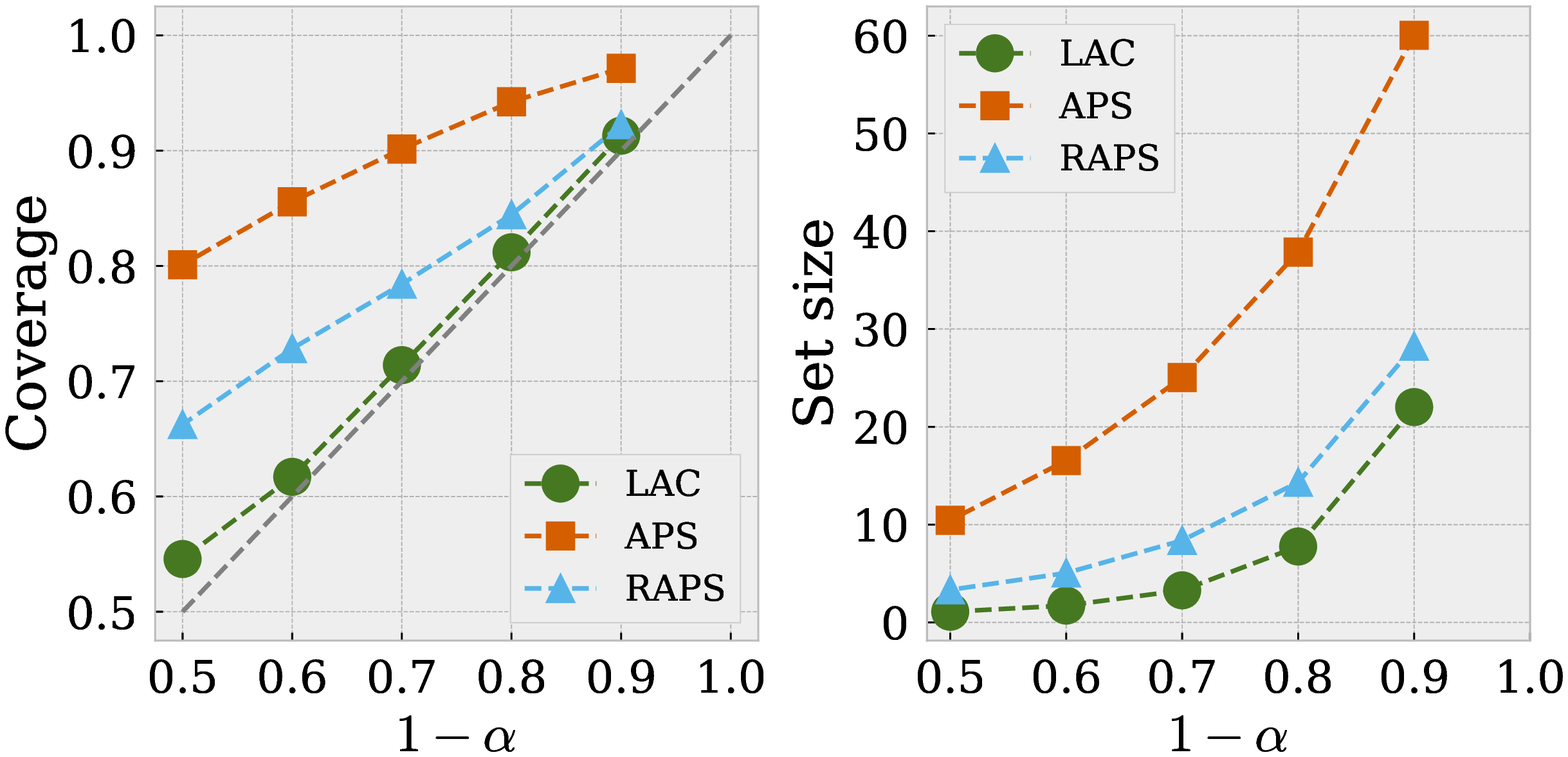}
      \caption{\textbf{Fitzpatrick17K}}
      % \label{fig:challenge-efficiency}
    \end{subfigure}
    \vspace{-0.20in}

    % \vskip 0.2in
    % \begin{center}
    % \centerline{\includegraphics[width=0.8\columnwidth]{figures/experiment-3-cifar100-comparing-score.eps}}
    \caption{\textbf{LAC is the most efficient conformal score function.} When disallowing empty prediction sets, adaptive score functions such as APS and RAPS have more coverage than necessary than the marginal guarantee. LAC has tighter coverage and a smaller average set size than APS and RAPS.}
    \label{fig:cifar-conformal-score-compare}
    \end{center}
    \vskip -0.2in
\end{figure}

\begin{figure}[ht] 
    \vskip 0.2in
    \begin{center}
      \centering
        \begin{subfigure}{.50\textwidth}
          \centering
          \includegraphics[width=\linewidth]{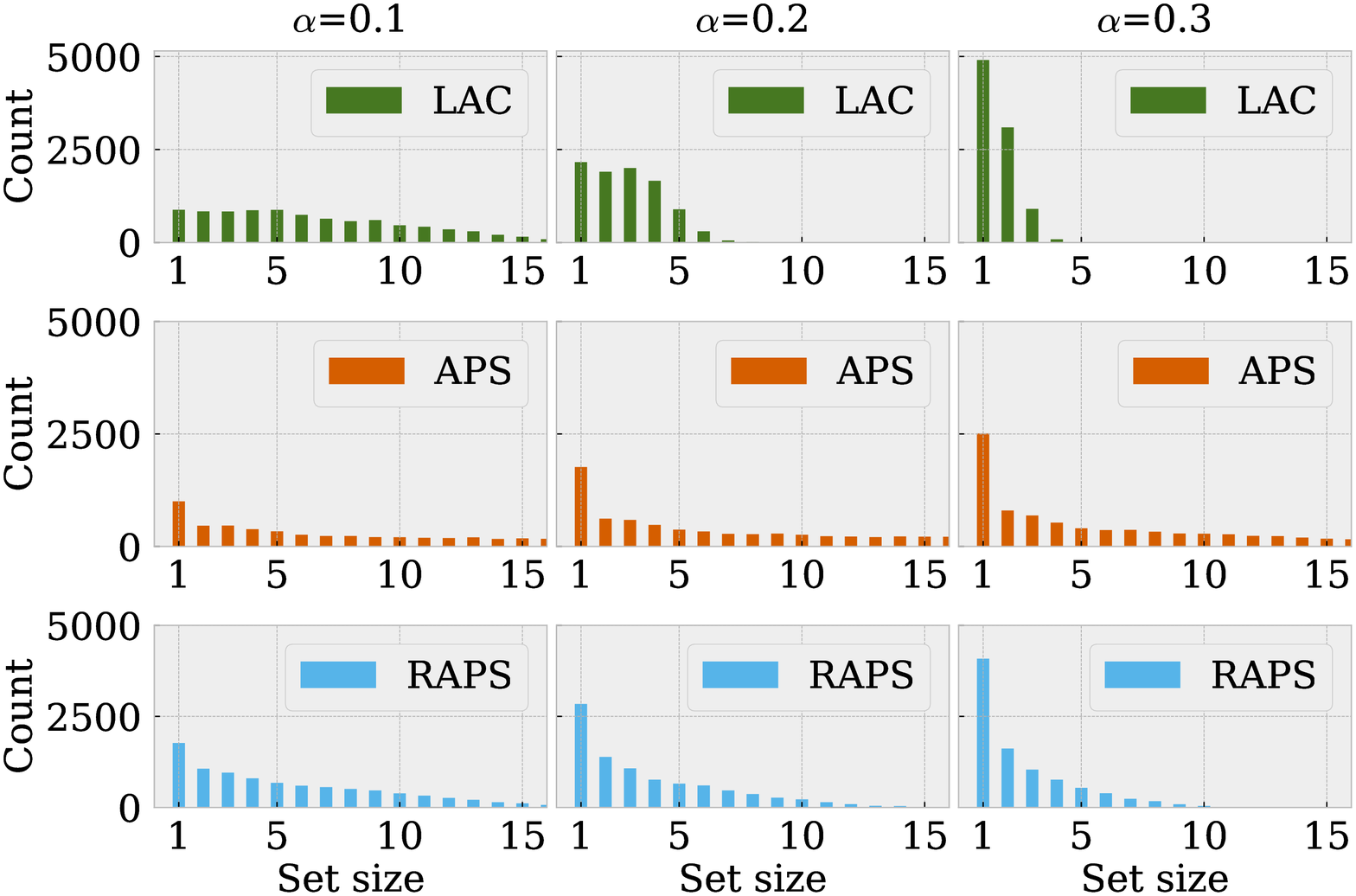}
          \caption{\textbf{CIFAR-100}}
          % \label{fig:challenge-exchangeability}
        \end{subfigure}\hfill
        \begin{subfigure}{.50\textwidth}
          \centering
          \includegraphics[width=\linewidth]{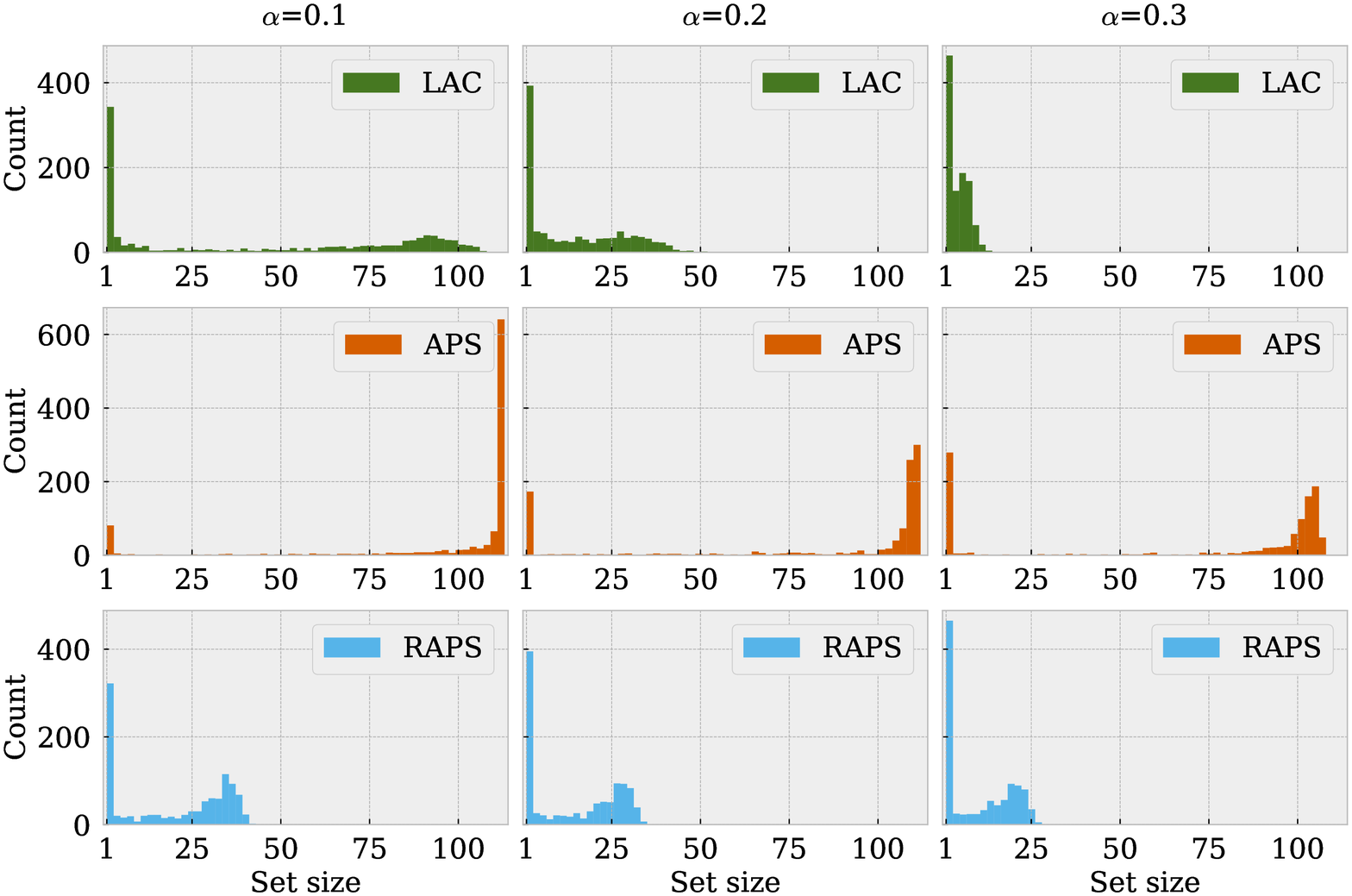}
          \caption{\textbf{Fitzpatrick17K}}
          % \label{fig:challenge-efficiency}
        \end{subfigure}
        \vspace{-0.20in}
    % \centerline{\includegraphics[width=0.6\columnwidth]{figures/cifar100-score_dist.eps}}
    \caption{\textbf{Comparing the distribution of prediction size between conformal score functions.} We trained a TCT model on CIFAR-100 with 20 heterogeneous clients and plotted the number of predictions at each set size. RAPS has a wider range of set sizes and a larger number of sets with a single element than LAC across $\alpha$ thresholds.}
    \label{fig:cifar-score-dist}
    \end{center}
    \vskip -0.2in
\end{figure}

\begin{figure*}[ht]
\vskip -0.2in
% \begin{center}
  \centering
\begin{subfigure}{.45\textwidth}
  \centering
  \includegraphics[width=\linewidth]{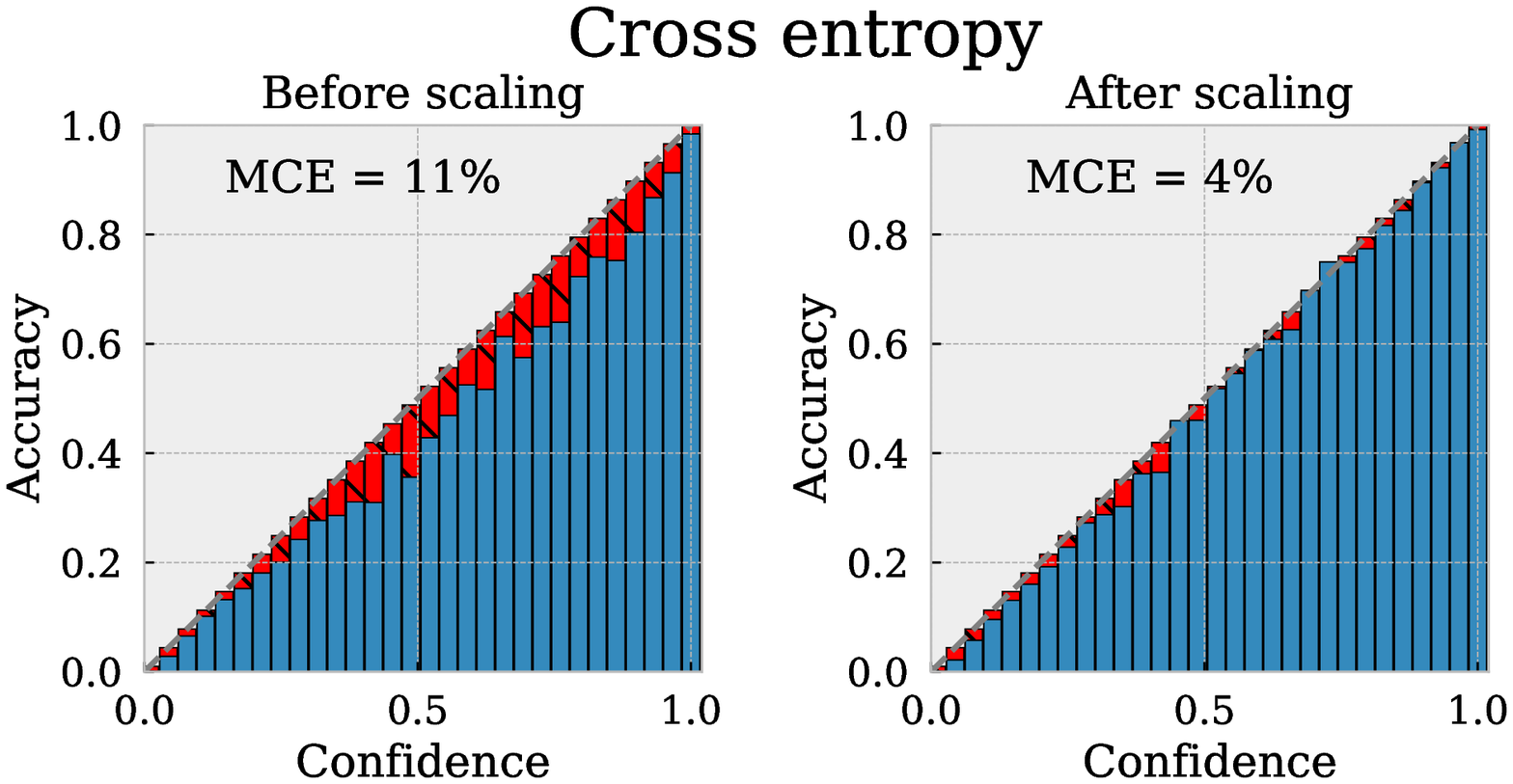}
  % \caption{A subfigure}
  % \label{fig:sub1}
\end{subfigure}%
\begin{subfigure}{.45\textwidth}
  \centering
  \includegraphics[width=\linewidth]{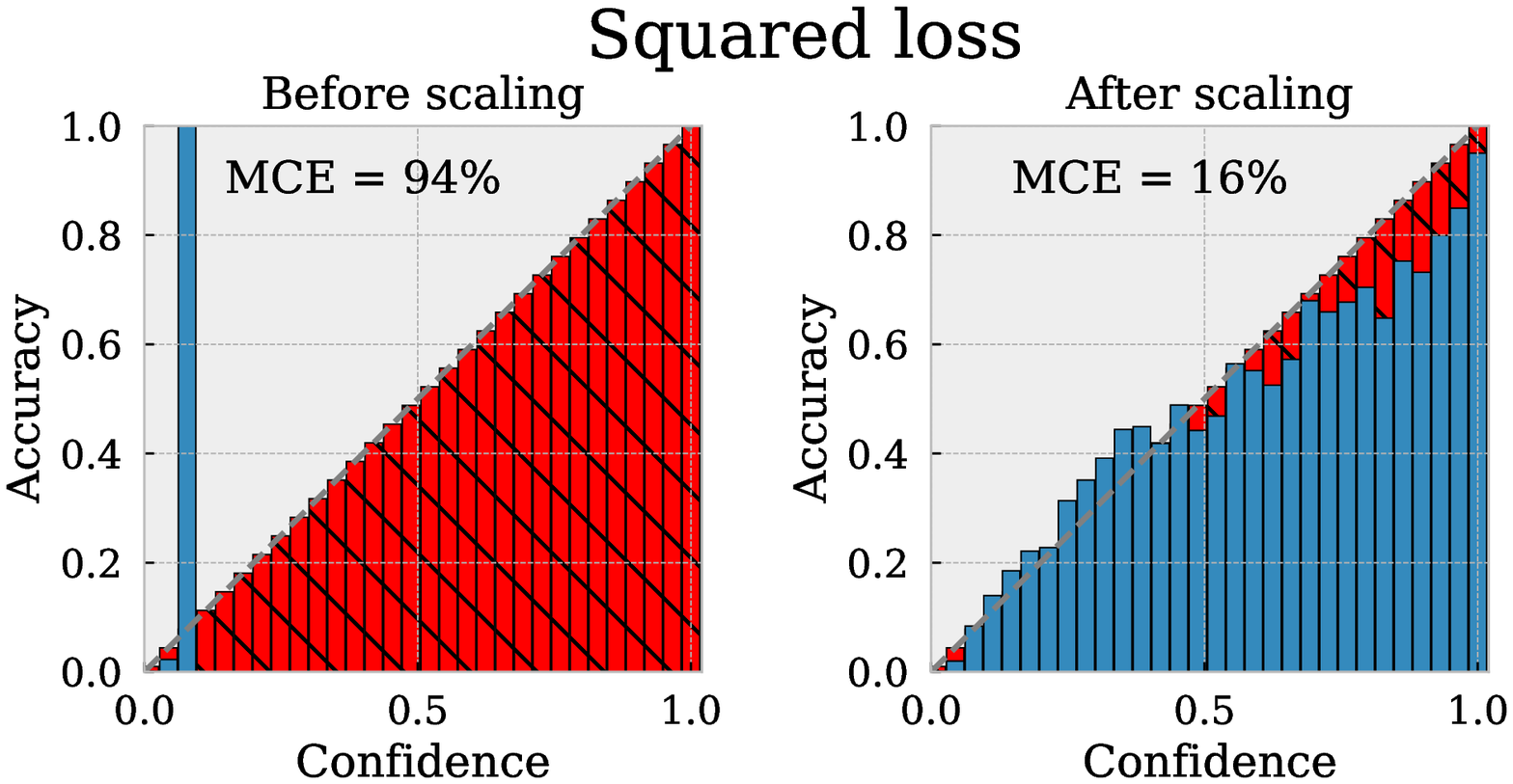}
  % \caption{A subfigure}
  % \label{fig:sub2}
\end{subfigure}
    \vspace{-0.2in}
\caption{\textbf{TCT trained with squared loss is poorly calibrated on CIFAR-100.} Comparing Maximum Calibration Error (MCE) of two loss functions TCT with IID partitions on the CIFAR-100 dataset. The red bins show ideal calibration, and the blue bins show observed calibration.}
\label{fig:loss-comparison-cifar100}
% \end{center}
\vskip -0.2in
\end{figure*}

\begin{figure*}[ht]
% \vskip 0.2in
% \begin{center}
  \centering
\begin{subfigure}{.45\textwidth}
  \centering
  \includegraphics[width=\linewidth]{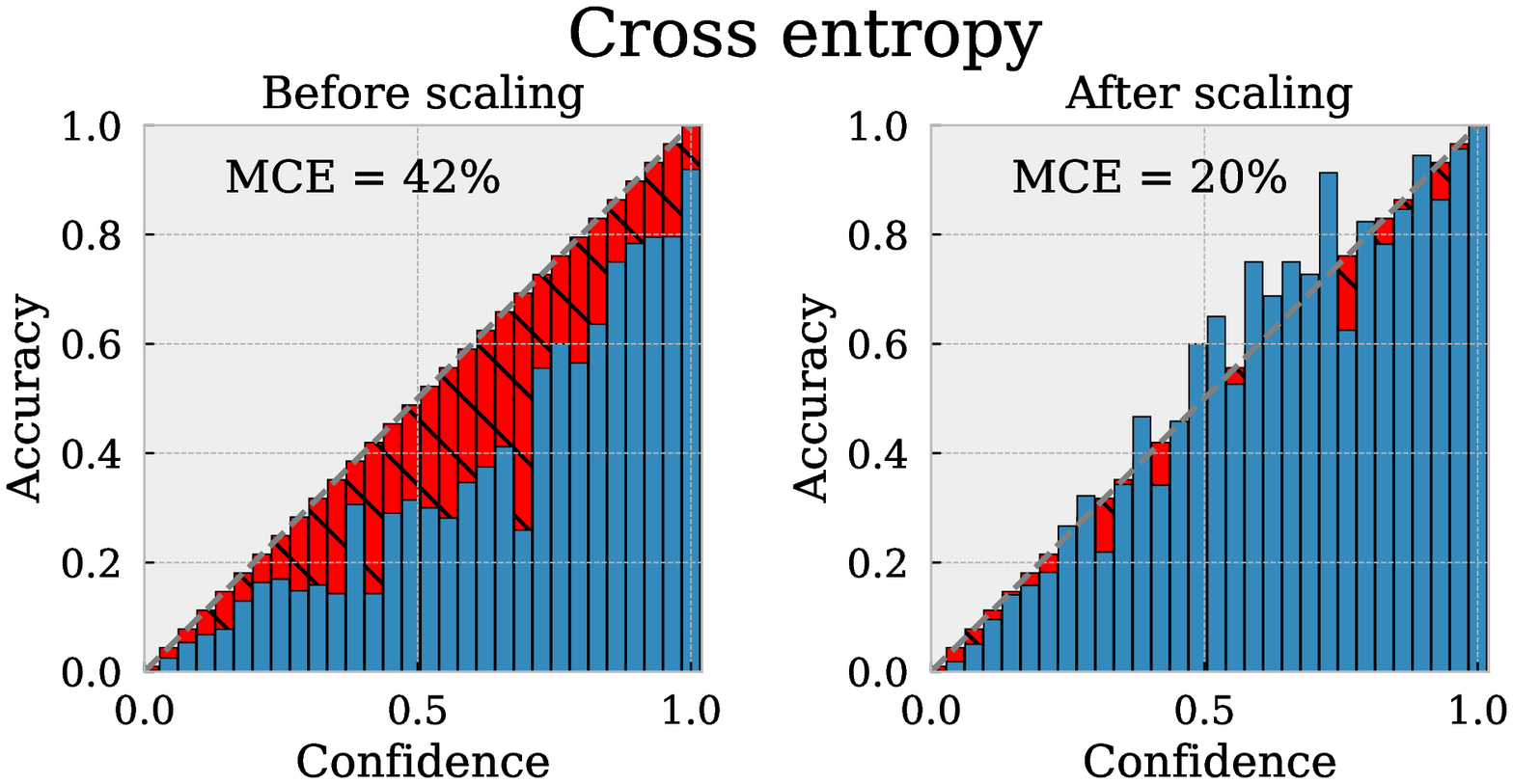}
  % \caption{A subfigure}
  % \label{fig:sub1}
\end{subfigure}%
\begin{subfigure}{.45\textwidth}
  \centering
  \includegraphics[width=\linewidth]{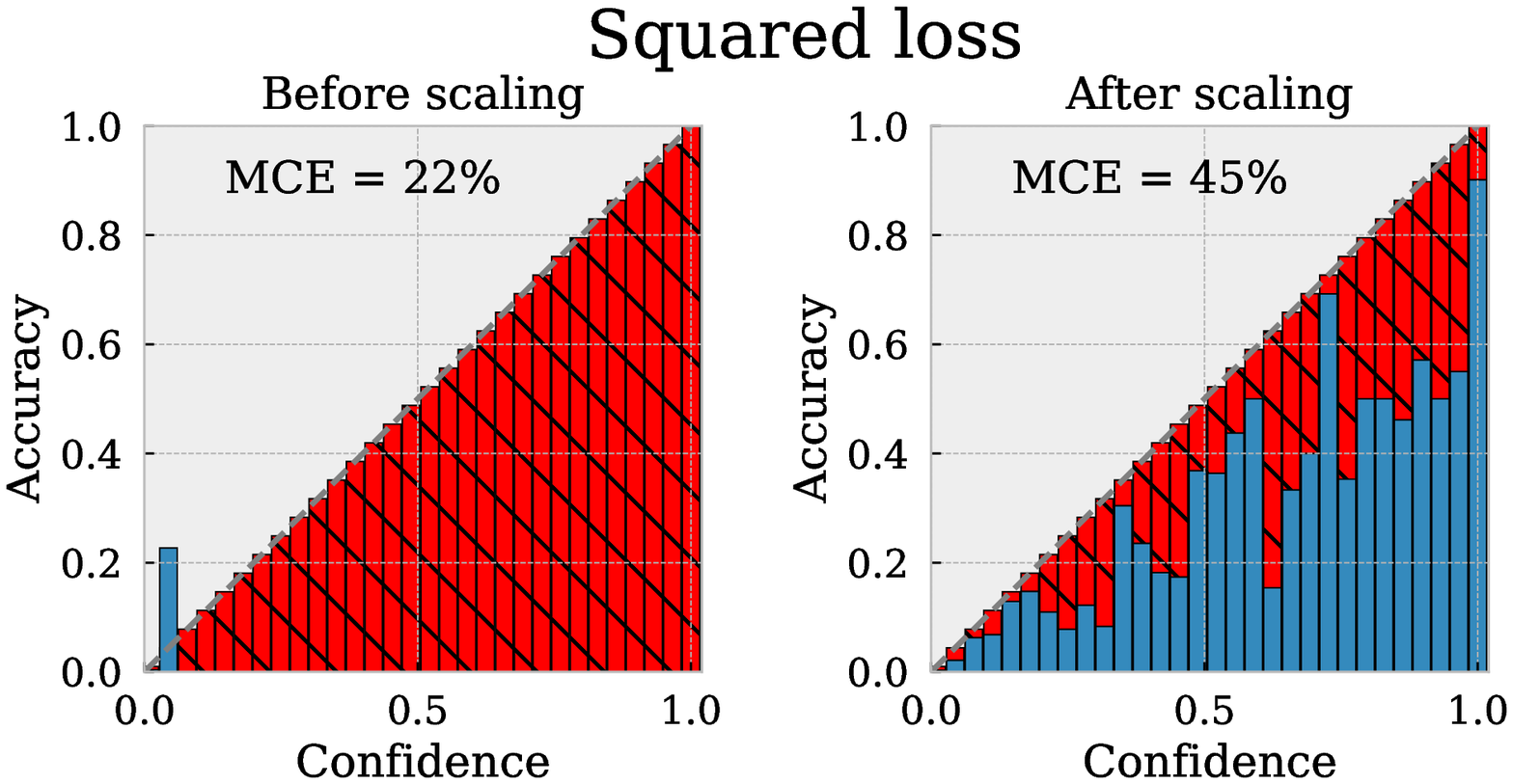}
  % \caption{A subfigure}
  % \label{fig:sub2}
\end{subfigure}
\caption{\textbf{TCT trained with squared loss is poorly calibrated on Fitzpatrick17k.} Comparing Maximum Calibration Error (MCE) of two loss functions TCT with IID partitions on Fitzpatrick17k dataset. The red bins show ideal calibration, and the blue bins show observed calibration.}
\label{fig:loss-comparison-fitzpatrick}
% \end{center}
\vskip -0.2in
\end{figure*}

\end{document}